\documentclass[runningheads]{llncs}
\usepackage{graphicx}
\usepackage{comment}
\usepackage{amsmath,amssymb} 
\usepackage{color}

\usepackage[width=122mm,left=12mm,paperwidth=146mm,height=193mm,top=12mm,paperheight=217mm]{geometry}

\usepackage{import}
\usepackage{caption}
\usepackage{epsfig}
\usepackage{multirow}
\usepackage[hidelinks]{hyperref}
\hypersetup{
	colorlinks   = true, 
	urlcolor     = blue, 
	linkcolor    = blue, 
	citecolor   = green 
}

\usepackage{booktabs}
\usepackage{makecell} 
\usepackage{breakcites} 

\DeclareMathOperator*{\argmax}{arg\,max}

\def\figwidth{14mm}

\iftrue
\newcommand{\kibok}[1]{\textcolor{red}{[Kibok: #1]}}
\newcommand{\zy}[1]{\textcolor{magenta}{Zhuoyuan: #1}}
\newcommand{\xcyan}[1]{\textcolor{blue}{Xinchen: #1}}
\newcommand{\ersin}[1]{\textcolor{cyan}{Ersin: #1}}
\else
\newcommand{\kibok}[1]{}
\newcommand{\zy}[1]{}
\newcommand{\xcyan}[1]{}
\newcommand{\ersin}[1]{}
\fi

\iftrue 
\newcommand{\cutsectionup}{\vspace*{-8pt}}
\newcommand{\cutsectiondown}{\vspace*{-4pt}}

\newcommand{\cutsubsectionup}{\vspace*{-8pt}}
\newcommand{\cutsubsectiondown}{\vspace*{-2pt}}

\newcommand{\cutparagraphup}{\vspace*{-2pt}}
\newcommand{\cutparagraphdown}{\vspace*{-0pt}}

\newcommand{\cutcaptionup}{\vspace*{-4pt}}
\newcommand{\cutcaptiondown}{\vspace*{-12pt}}

\newcommand{\cuttablecaptionup}{\vspace*{-0pt}}

\newcommand{\cuttableup}{\vspace*{-0pt}}
\newcommand{\cuttabledown}{\vspace*{-12pt}}

\newcommand{\cutabstractup}{\vspace*{-10pt}}
\newcommand{\cutabstractdown}{\vspace*{-20pt}}

\else 
\newcommand{\cutsectionup}{}
\newcommand{\cutsectiondown}{}

\newcommand{\cutsubsectionup}{}
\newcommand{\cutsubsectiondown}{}

\newcommand{\cutparagraphup}{}
\newcommand{\cutparagraphdown}{}

\newcommand{\cutcaptionup}{}
\newcommand{\cutcaptiondown}{}

\newcommand{\cuttablecaptionup}{}

\newcommand{\cuttableup}{}
\newcommand{\cuttabledown}{}

\newcommand{\cutabstractup}{}
\newcommand{\cutabstractdown}{}

\fi

\begin{document}
	\pagestyle{headings}
	\mainmatter
	
	\def\ECCVSubNumber{2349}  
	
	\title{ShapeAdv: Generating Shape-Aware\\Adversarial 3D Point Clouds}
	
	
	%
	\author{
		Kibok Lee\thanks{Work done during internship with Uber ATG.}$^{1,2}$\quad
		Zhuoyuan Chen$^{1}$\quad
		Xinchen Yan$^{1}$\\
		Raquel Urtasun$^{1,3}$\quad
		Ersin Yumer$^{1}$\\
	}
	\institute{
		$^{1}$Uber ATG\quad
		$^{2}$University of Michigan\quad
		$^{3}$University of Toronto\\
		$^{1}${\tt\small \{zhuoyuan, xcyan, urtasun, yumer\}@uber.com} \quad
		$^{2}${\tt\small kibok@umich.edu}
	}
	%
	%
	\maketitle
	

\cutabstractup
\begin{abstract}
%
We introduce ShapeAdv, a novel framework to study shape-aware adversarial perturbations that reflect the underlying shape variations (e.g., geometric deformations and structural differences) in the 3D point cloud space.
We develop shape-aware adversarial 3D point cloud attacks by leveraging the learned latent space of a point cloud auto-encoder where the adversarial noise is applied in the latent space.
Specifically, we propose three different variants including an exemplar-based one by guiding the shape deformation with auxiliary data, such that the generated point cloud resembles the shape morphing between objects in the same category.
Different from prior works, the resulting adversarial 3D point clouds reflect the shape variations in the 3D point cloud space while still being close to the original one.
In addition, experimental evaluations on the ModelNet40 benchmark demonstrate that our adversaries are more difficult to defend with existing point cloud defense methods and exhibit a higher attack transferability across classifiers.
Our shape-aware adversarial attacks are orthogonal to existing point cloud based attacks and shed light on the vulnerability of 3D deep neural networks.
\end{abstract}
\cutabstractdown


\cutsectionup
\section{Introduction} \label{sec:intro}
\cutsectiondown

\begin{figure}[h]
\centering
\includegraphics[width=0.8\linewidth]{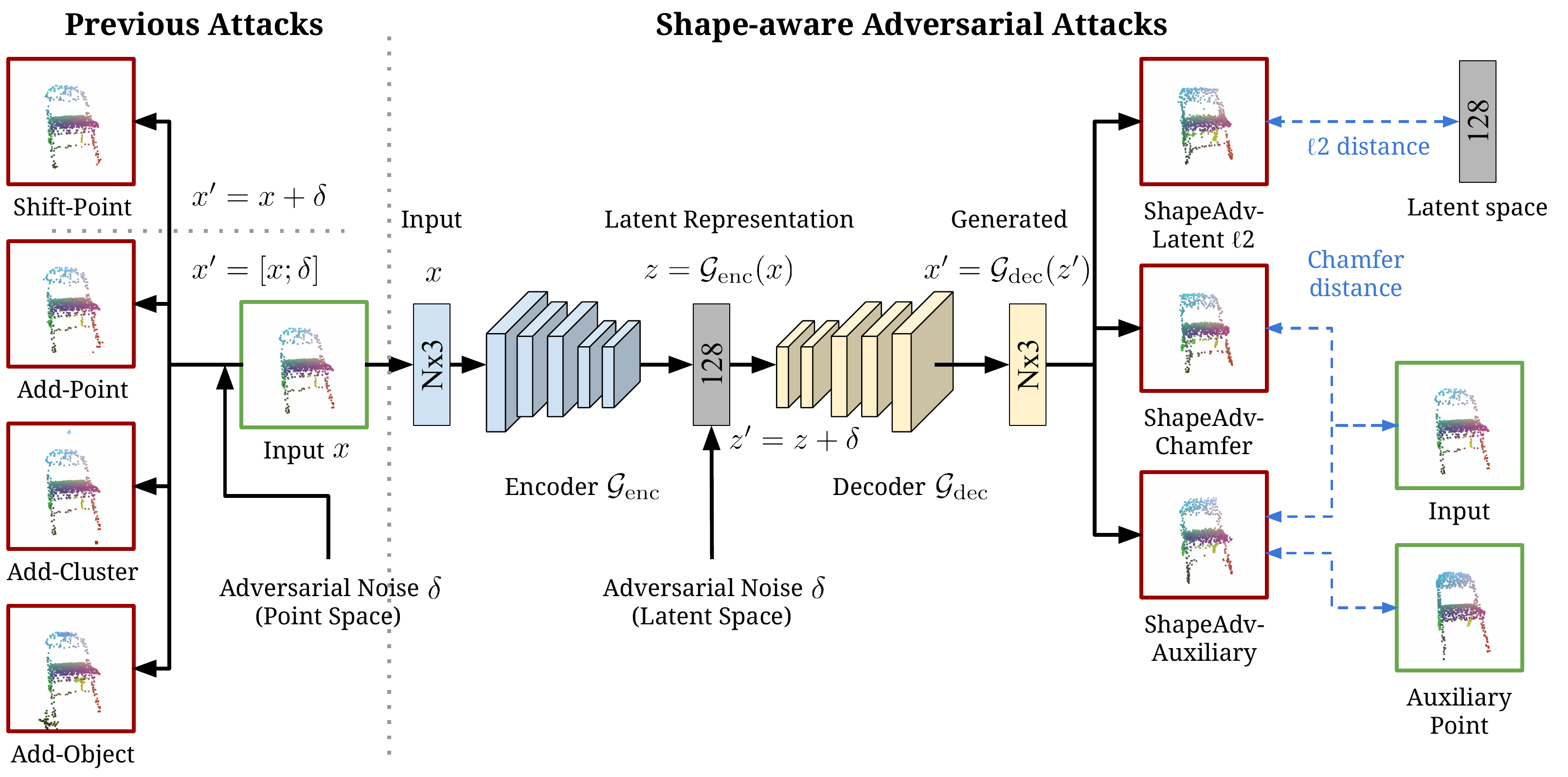}
\cutcaptionup
\captionof{figure}{
An overview of the proposed \textit{shape-aware} adversarial 3D point cloud generation.
We propose a novel framework to inject adversarial noise into 3D point cloud by leveraging the latent space of a point cloud auto-encoder.
Our method first learns the latent representations with a point cloud auto-encoder in an unsupervised way through point reconstruction.
Second, we propose three different methods (ShapeAdv-Latent $\ell_2$, ShapeAdv-Chamfer and ShapeAdv-Auxiliary) to 
inject perturbations in the latent space and generate the adversarial 3D point cloud using the decoder network.
Our ShapeAdv is orthogonal to existing attack methods including shifting points, adding independent point perturbations, adding clusters and objects.
}
\cutcaptiondown
\label{fig:intro}
\end{figure}

Deep neural networks (DNNs)~\cite{qi2017pointnet,qi2017pointnet++,wang2018dynamic} have achieved state-of-the-art 
performance in various 3D perception tasks~\cite{brock2016generative,landrieu2018large,qi2018frustum,qi2016volumetric,sedaghat2017orientation,zhou2018voxelnet} 
and have been widely applied in safety-critical applications including robotic grasping~\cite{mahler2017dex,yan2018deep} 
and autonomous driving~\cite{shi2019pointrcnn,yang2018pixor}.
However, several recent studies~\cite{cao2019adversarial,su2018deeper,xiang2019generating,xiao2019meshadv} 
raised concerns regarding the vulnerability of DNNs when applied to 3D sensory data such as point clouds.
For example, carefully crafted adversarial 3D point clouds can induce arbitrary prediction errors in modern platforms.
%
Recent works~\cite{liu2019extending,xiang2019generating,yang2019adversarial} proposed several heuristics to construct adversarial 3D point clouds by injecting adversarial noise directly to the victim object.
Among the heuristics, attacks based on independent point shifting or addition can be resolved by statistical denoising 
or outlier removal mechanisms~\cite{zhou2019dup}, while attacks based on adding adversarial clusters 
or objects fail to capture shape variations of a certain 3D point cloud due to significant changes 
to the true shape of the victim object.
Indeed, modeling shape variations of a 3D point cloud is challenging as it requires reasoning about the \textit{structural} factors (e.g., two chair legs are connected by a horizontal bar), \textit{geometric} factors (e.g., length of the chair leg), and the interactions in between.
At the same time, understanding the failure modes of DNNs towards adversarial 3D perturbations which reflect shape variations in the point cloud space is an important but under-explored.

In this work, we are interested in generating \textit{shape-aware} adversarial point perturbations to an existing 3D point cloud. 
More specifically, we would like our perturbations to reflect certain shape variations in local geometry, global geometry, or structures while keeping the overall shape close to the original 3D point cloud.
We propose a two-stage framework to generate \textit{shape-aware} adversarial examples or \textit{ShapeAdv} by leveraging a widely used point cloud auto-encoder network~\cite{achlioptas2018learning,groueix2018atlasnet,yang2019pointflow}.
First, we learn to encode the high-dimensional 3D point cloud data into a lower-dimensional compact representation in the latent space through unsupervised shape reconstruction.
As the auto-encoder network is well trained to reconstruct input shape, the learned latent space contains essential geometry and structure information about the 3D object shape that approximates the \textit{shape manifold}. 
Second, we propose to add perturbation in the latent space and use our decoder network to propagate the signal to the 3D point cloud space.
Consequently, a small perturbation on the latent space would cause a small change on the point space, so that the resulting point cloud is still close to the input shape.
This is different from the previous approaches which directly add an adversarial noise to the point space.
Our method is also supported by the empirical finding~\cite{achlioptas2018learning,girdhar2016learning,wu2016learning,yang2019pointflow} that linear interpolation on the learned latent space can produce smooth and continuous shape morphing.

Furthermore, we extend the proposed framework by incorporating auxiliary point clouds from the same category as the input when generating adversarial attacks.
As multiple shapes are used to restrict the shape of the adversary, the generated adversarial shape resembles both the input and auxiliary point clouds.
%
%
Finally, we demonstrate that auto-encoders can also be used for adversarial defense for 3D point clouds.
%
%

To show the effectiveness of our method, we attack the commonly used PointNet classifier~\cite{qi2017pointnet} on the ModelNet40~\cite{sedaghat2017orientation,wu20153d} dataset.
We compare our proposed attacks with previous methods:
adding or shifting points, and adding clusters or objects~\cite{xiang2019generating}.
While all attacks can achieve $100\%$ attack success rate, we observe that our attacks are hard to defend against compared to simple baseline attacks which introduce only a few outliers;
on the other hand, adding clusters or objects significantly change the geometry of the input. 
We experiment with two different auto-encoder networks, namely, the MLP baseline~\cite{achlioptas2018learning} and AtlasNet~\cite{groueix2018atlasnet}.
Our results show that AtlasNet preserves the detail of shapes as it generates adversarial point clouds by deforming the local surfaces which can be partially removed by existing defense methods.
In comparison, the MLP baseline generates a global geometric deformation as the adversarial perturbation, which makes the defense challenging.
We consider this as a trade-off between the expressive power of point decoder and the strength of adversarial perturbations.
%
%
%
Finally, we show that our proposed attacks improve black-box transferability to other models, such as PointNet++~\cite{qi2017pointnet++} and DGCNN~\cite{wang2018dynamic}.

To summarize, the contributions of our shape-aware adversarial 3D point cloud generation are as follows:
\begin{itemize}
\item We introduce a novel framework for generating shape-aware adversarial 3D point clouds by injecting an adversarial noise in the latent space of a point auto-encoder.
\item We propose three different variants to generate such shape-aware attacks with different additional constraint:
by minimizing the difference between the input and adversary in either the point space or latent space, and also by incorporating auxiliary point clouds from the same category with the input in the constraint.
%
%
\item We illustrate that the injected shape-aware perturbation reflects certain shape variations (e.g., \textit{geometric} deformation or \textit{structural} difference) in the 3D point cloud space.
\item Compared to existing point cloud attack methods that directly operate in the raw point cloud space, 
the proposed attack methods are significantly more difficult to defend against with existing defense methods and exhibit higher attack transferability.
%
%

%
%
\end{itemize}

\vspace*{-10pt}


\cutsectionup
\section{Related work}
\cutsectiondown

\cutparagraphup
\paragraph{Deep 3D shape generation.}
\cutparagraphdown
Generative 3D shape modeling~\cite{blanz1999morphable,bogo2016keep} has become a popular 
research topic in machine learning, computer vision, and graphics.
The past few years have witnessed tremendous
advances in
deep generative modeling of 3D voxels~\cite{choy20163d,girdhar2016learning,wu2017marrnet,wu2016learning,wu20153d,yan2016perspective},
point clouds~\cite{achlioptas2018learning,fan2017point,gadelha2018multiresolution,groueix2018atlasnet,jiang2018gal,mandikal20183d,yang2019pointflow,yang2018foldingnet}, and
surface meshes~\cite{gkioxari2019mesh,kanazawa2018end,li2017grass,liu2019soft,tewari2017mofa,tung2017self,varol2018bodynet,zuffi2019three}. 
%
Wu et al.~\cite{wu20153d} introduced the first deep generative models on 3D volumetric shapes using stacked 
Restricted Boltzmann Machines with 3D convolutions operating on 3D occupancy grids.
%
\cite{girdhar2016learning,wu2017marrnet,wu2016learning,yan2016perspective} extended this framework with encoder-decoder 3D convolutional architectures and in-network LSTM modules~\cite{choy20163d}.

Due to several practical challenges (e.g., memory and computationally expensive 3D convolution operations) in generating high-fidelity 3D voxels, most recent works have shifted the focus to generative modeling of point clouds.
Compared to the voxel representation, operations in point clouds are computationally more efficient;
for example, the dimension of voxel fits 3D convolution, while that of point clouds fits 1D convolution.
However, point clouds require permutation-invariant operations because the order of points can be random, which limits a straightforward transfer from other domains.
%
Fan et al.~\cite{fan2017point} introduced a deep learning-based framework to synthesize 3D point cloud from a single image using the Chamfer distance and Earth mover's distance to better preserve the shape invariance under point permutation.
Achlioptas et al.~\cite{achlioptas2018learning} proposed a two-step training method for an auto-encoder, for learning latent representation space of 3D point clouds and generating 3D point clouds from the latent representation.
To improve the MLP based point auto-encoder, Groueix et al.~\cite{groueix2018atlasnet} generates 3D point clouds in a piece-wise planar fashion where each planar surface is deformed by a separate neural network.
%
Our proposed shape-aware adversarial method can be crafted with different architectures, as point auto-encoder is a widely used architecture for shape reconstruction and generation.
Besides 3D point cloud representation, the idea of generating an adversarial attack on the latent space can potentially be applicable to other 3D data representations, such as voxel grids and surface meshes.

\cutparagraphup
\paragraph{Adversarial examples.}
\cutparagraphdown
Early studies~\cite{goodfellow2014explaining,szegedy2013intriguing} have revealed the vulnerability of modern image classifiers by carefully crafted adversarial examples, which can induce arbitrary prediction errors while being imperceptible by human.
This is achieved by optimizing an objective that maximizes the prediction errors while restricting the perturbation magnitude under an $\ell_p$ norm.
Adversarial attacks have gained tremendous attentions in machine learning, computer vision, and security communities with two parallel efforts focused on generating adversarial attacks~\cite{carlini2017towards,moosavi2016deepfool,papernot2017practical,papernot2016limitations,xiao2018generating,zhao2018natural} and devising effective defenses against
such mechanisms~\cite{carlini2016defensive,jalal2017robust,jin2019ape,meng2017magnet,ranjan2017improving,samangouei2018defense,schott2019towards,xu2017feature,yan2018deep} on natural images.

While adversarial examples in the 2D image domain have been extensively studied, generating 3D adversarial examples is relatively under-explored.
Xiao et al.~\cite{xiao2019meshadv} introduced a method to inject adversarial 3D shape deformations 
to a victim object so as to fool the 2D object detectors when projecting into the 2D image space using a differentiable renderer.
Cao et al.~\cite{cao2019adversarial} further investigated machine learning-based LiDAR spoofing attacks in the real world.
Recently, several works~\cite{liu2019extending,su2018deeper,xiang2019generating,yang2019adversarial} have investigated the problem of generating adversarial signals on the raw 3D point clouds.
%
%
For example, among the attacks proposed by \cite{xiang2019generating},
adding or shifting few points is easily defended by outlier removal methods~\cite{liu2019extending,zhou2019dup}, and
adding a few clusters or objects makes a significant difference from the original shape, such that an appropriate preprocessing or segmentation stage can remove them.

Unlike prior works that generate adversarial noises directly in the point cloud space, 
our work injects adversarial perturbations on the learned \textit{shape manifold} 
using latent representation of a point auto-encoder, 
such that the adversary reflects the shape variations of a 3D point cloud.
Our work is also related to recent studies on semantic or ``unrestricted'' adversarial 
examples in the image domain~\cite{alzantot2018generating,bhattad2019big,frosst2018darccc,joshi2019semantic,qiu2019semanticadv,song2018constructing,stutz2019disentangling} using generative models.
To the best of our knowledge, the proposed shape-aware adversarial point cloud generation framework 
is the first study on 3D point cloud adversaries in a shape-aware fashion using the latent space 
of a point cloud auto-encoder.




\cutsectionup
\section{ShapeAdv: Shape-Aware Adversarial Attacks}
\cutsectiondown
\label{sec:method}

In this section, we describe our proposed ShapeAdv attacks and defense methods.
%

\cutsubsectionup
\subsection{Background}
\label{sec:background}
\cutsubsectiondown


\cutparagraphup
\paragraph{Adversarial attacks on 3D point clouds.}
\cutparagraphdown
A 3D point cloud $x \in \mathcal{X} \subset \bigcup_{n=1}^{\infty} \mathbb{R}^{n \times 3}$ 
is a set of $N$ points sampled from the surface of an object, 
%
where each point is represented by a tuple of Cartesian coordinates (XYZ) in the point space $\mathcal{X}$.
Let $\mathcal{M}: \mathcal{X} \rightarrow \mathcal{Y}$ be a classification model, 
which takes a 3D point cloud $x$ as an input and predicts its label $y \in \mathcal{Y} \subset \mathbb{Z}$.
Since the number of points $N$ and their order vary depending on the sampling, $\mathcal{M}$ is designed to be invariant to the dimension of $N$, usually achieved by a global \texttt{max pool} or \texttt{average pool} operation~\cite{qi2017pointnet,wang2018dynamic}.

Given a point cloud data and its ground truth label $(x,y)$, the goal of an adversarial attack is to lead $\mathcal{M}$ to misclassify the label of an input $x$, by finding an adversary $x'$ satisfying:
\begin{align}
\min_{x'} \mathcal{D}(x, x') \quad \text{s.t.} \quad \mathcal{M}(x') \neq y,
\label{eqn:adv_x}
\end{align}
where $\mathcal{D}$ is a distance metric to measure the similarity between the original data $x$ and the optimized adversary $x'$.
In targeted attacks, a target label $t \neq y$ is provided such that the adversary is optimized to satisfy $\mathcal{M}(x') = t$, while $\mathcal{M}(x') \neq y$ is sufficient in untargeted attacks.

In adversarial attacks on 3D point clouds, two types of strategies
have been studied~\cite{liu2019extending,su2018deeper,xiang2019generating,yang2019adversarial}:
adversarial point perturbation and adversarial point generation.
Adversarial point perturbation adds a small perturbation $\delta$ to $x$, i.e., $x' = x + \delta$,
%
while adversarial point generation augments $N_\text{aug}$ points to $x$, i.e., $x' = [x; \delta] \in \mathbb{R}^{(N+N_\text{aug}) \times 3}$.
%

\cutparagraphup
\paragraph{Learned latent space as an approximation to the shape manifold.}
\cutparagraphdown
Different from prior works, we propose to inject an adversarial perturbation to the latent representation of a point cloud.
To achieve this, we take an auto-encoder $\mathcal{G} = \mathcal{G}_\text{dec} \circ \mathcal{G}_\text{enc}$, which consists of an encoder network $\mathcal{G}_\text{enc}$ and a decoder network $\mathcal{G}_\text{dec}$.
In this setting, given a data and its label $(x,y)$,
we optimize the adversary in the latent space $z'$ to be similar to the encoded data $z = \mathcal{G}_\text{enc}(x)$, such that it leads the model $\mathcal{M}$ to misclassify the decoded output $x' = \mathcal{G}_\text{dec}(z')$.
Similar to Eq.~\eqref{eqn:adv_x}, we optimize the following:
\begin{align}
\min_{z'} \mathcal{D}(z, z') \quad \text{s.t.} \quad \mathcal{M}(x') \neq y,
\label{eqn:adv_z}
\end{align}
where $\mathcal{D}$ is a distance metric to measure the similarity between the input and its adversary in the latent space.
For this purpose, any auto-encoder is applicable, but a high-quality auto-encoder would produce better quality of the generated adversarial attack, i.e., $\mathcal{D}(x, \mathcal{G}(x))$ should be small enough for any $x \in \mathcal{X}$.

\cutparagraphup
\subsection{Shape-Aware Adversarial Attacks in the Latent Space}
\label{sec:attack}
\cutparagraphdown
Since Eq.~\eqref{eqn:adv_z} could not be directly optimized via gradient descent, 
we reformulate the optimization problem as:
\begin{align}
\min_{z'} \mathcal{L}_\text{adv}(y, \mathcal{M}(x')) + \lambda \mathcal{D}(z, z'),
\label{eqn:shapeadv}
\end{align}
where $\lambda$ controls the balance between the adversarial loss and the similarity between the input and the adversary,
$\mathcal{D}$ is the distance metric,
$z = \mathcal{G}_\text{enc}(x)$ is the latent representation of the input data, and
$x' = \mathcal{G}_\text{dec}(z')$ is the adversary in the point space $\mathcal{X}$.
Here, any adversarial loss can be applicable, but we follow the most effective formulation of the CW attack~\cite{carlini2017towards}, as described below:
let $l(x)_{y'}$ be the logits (or unnormalized probabilities) predicted by the model $\mathcal{M}$ that $x$ 
is from the label $y'$, such that $\mathcal{M}(x) = \argmax_{y'} l(x)_{y'}$. 
%
Then, for a targeted attack,
\begin{align}
\mathcal{L}_\text{adv}^\text{T}(y, \mathcal{M}(x')) =
(\max_{y' \neq t} l(x')_{y'} - l(x')_t)^+,
\label{eqn:shapeadv_targeted}
\end{align}
and for an untargeted attack,
\begin{align}
\mathcal{L}_\text{adv}^\text{U}(y, \mathcal{M}(x')) &=
(l(x')_y - \max_{y' \neq y} l(x')_{y'})^+,
\label{eqn:shapeadv_untargeted}
\end{align}
where $(\cdot)^+ = \max(\cdot, 0)$ is the hinge loss.
In the following, we specify different choices of the regularization $\mathcal{D}$ in our approach. 

\cutparagraphup
\paragraph{Shape-aware attack in the latent space.}
\cutparagraphdown
%
It is straightforward to minimize the difference between the input and the adversary in the latent space.
We then solve the following optimization problem:
\begin{align}
\min_{z'} \mathcal{L}_\text{adv}(y, \mathcal{M}(x')) + \lambda \lVert z - z' \rVert_2,
\label{eqn:shapeadv_l2}
\end{align}
where $x' = \mathcal{G}_\text{dec}(z')$ is the decoded adversary.
We regularize $z$ and $z'$ to be close with respect to $\ell_2$ distance.
%
To distinguish with other proposed methods, we refer to this method as \textit{ShapeAdv-Latent $\ell_2$}.

\cutparagraphup
\paragraph{Shape-aware attack in the point space.}
\cutparagraphdown
In ShapeAdv-Latent $\ell_2$, we implicitly assume that local similarity is preserved:
i.e., as long as $z'$ is similar to $z$, the adversary $x'$ is also similar to $x$.
%
%
Alternatively, we can directly constrain the similarity between the decoded adversary $x'$ and the input $x$, 
as our point decoder $\mathcal{G}_\text{dec}$ is differentiable.
More specifically, let $x = [x_1, x_2, \dots, x_N]^\top$ and $x' = [x_1', x_2', \dots, x_{N'}']^\top$ be the collection of points in the input and adversary, respectively.
We apply the squared Chamfer distance $\mathcal{D}_\text{CH}^2(x, x') = \frac{1}{\lVert x' \rVert_0} \sum_{x_j'} \min_{x_i} \lVert x_i - x_j' \rVert_2^2 + \frac{1}{\lVert x \rVert_0} \sum_{x_i} \min_{x_j'} \lVert x_i - x_j' \rVert_2^2$ as the metric,
which is permutation-invariant and effective in approximating the shape manifold of 3D point cloud data~\cite{fan2017point}:\footnote{Strictly speaking, the Chamfer distance is not a valid metric because it does not satisfy the triangle inequality.
However, it is empirically shown to be effective.}
\begin{align}
\min_{z'} \mathcal{L}_\text{adv}(y, \mathcal{M}(x')) + \lambda \mathcal{D}_\text{CH}^2 (x, x'),
\label{eqn:shapeadv_ch}
\end{align}
where $x' = \mathcal{G}_\text{dec}(z') \in \mathcal{X}$ is the decoded adversary.
We refer to this method as \textit{ShapeAdv-Chamfer}.
%
%

\cutparagraphup
\paragraph{Shape-aware attack with auxiliary point clouds.}
\cutparagraphdown
In the above two methods, we optimize the adversarial perturbation in any direction in the latent space.
However, if the quality of the latent space is not perfect, the perturbation 
in the latent space would cause an undesirable perturbation in the point space.
%
To avoid such perturbation in the point space, we propose to leverage auxiliary 
point clouds sampled from the category of the input to guide the direction in the latent space.
%
Specifically, given a pair of data and its label $(x,y)$, we find $K$ nearest training data whose label is $y$.\footnote{We use the Chamfer distance in the point space to find the $K$ nearest neighbors.}
Let $x_0 = x$ be the input data and its nearest neighbor training data be $\{x_1, x_2, \cdots, x_K\}$.
We then solve the following optimization problem:
\begin{align}
\min_{z'} \mathcal{L}_\text{adv}(y, \mathcal{M}(x')) + \frac{\lambda}{K+1} \sum_{k=0}^{K} \mathcal{D}_\text{CH}^2 (x_k, x'),
\label{eqn:shapeadv_aux}
\end{align}
where $x' = \mathcal{G}_\text{dec}(z') \in \mathcal{X}$ is the decoded adversary.
The coefficient $1/(K+1)$ in the second term could be a tunable hyperparameter 
to give different weights to the input and auxiliary point clouds, 
but we could not find significant differences in terms of the attack performance quantitatively.
We refer to this method as \textit{ShapeAdv-Auxiliary}.
%
%
%
%

To analyze the adversary $x'$ generated by this attack, suppose we have only one auxiliary point cloud.
%
Then minimizing the second term in Eq.~\eqref{eqn:shapeadv_aux} forces $x'$ to be similar 
to both the input data $x$ and the auxiliary point cloud $x_1$, and its optimal value 
is expected to be a linear interpolation between them.
We note that this holds in a valid metric only.
Since the Chamfer distance is not a valid metric, the triangle inequality does not hold in general, 
such that the optimal value may not be a linear interpolation of $x$ and $x'$.
However, we use the Chamfer distance here, because we found that its qualitative results 
empirically follow our analysis and show a good performance.
In general, when we take $K$ auxiliary point clouds, the adversary $x'$ would expected to be 
in a convex polyhedron $\Omega(x_0, x_1, x_2, ..., x_K)$.
Therefore, this formulation also serves as a method to analyze the robustness of the classifier under shape deformation in the same class;
in other words, if such an adversary exists, then the model is not robust to shape deformation.
%

On the other hand, we can restrict the search space of adversary in the latent space $z'$ to be a convex polyhedron $\Omega(z_0, z_1, z_2, ...,z_K)$.
In this case, we no longer need the second term in Eq.~\eqref{eqn:shapeadv_aux} and $z'$ is formulated as a convex combination of the input and the auxiliary point clouds, i.e., $z' = \sum_{k=0}^{K} \alpha_k z_k$ where $\alpha_k$ is optimizable.
%
%
However, we empirically found that such formulation is not very effective.

\cutsubsectionup
\subsection{Shape-Aware Adversarial Defense}
\label{sec:defense}
\cutsubsectiondown


While we focused on using point auto-encoders to generate attacks, they can also be used to defend against adversarial point clouds.
First, as our latent space approximates the shape manifold, adversarial attacks out of the shape manifold can be projected on the shape manifold when it is encoded by $\mathcal{G}_\text{enc}$, such that the output decoded by $\mathcal{G}_\text{dec}$ may not have an adversarial perturbation;
in other words, auto-encoding has an effect of noise removal,
if the noise is out of the shape manifold.
Therefore, given a test input $x^\text{test}$ which can be either a clean or adversarial data, $\hat{x}^\text{test} = \mathcal{G}(x^\text{test})$ is the defended output, such that the prediction after defense $\hat{y}^\text{test}$ can be expressed as
\begin{align}
\hat{y}^\text{test} = (\mathcal{M} \circ \mathcal{G})(x^\text{test}).
\label{eqn:ae}
\end{align}

Similar defense mechanisms have also been studied in recent works~\cite{jalal2017robust,jin2019ape,samangouei2018defense,schott2019towards,song2018pixeldefend,stutz2019disentangling}, which mostly focused on evaluating their method in a small 2D image dataset, such as MNIST.
In contrast to the 2D image representation, the 3D point cloud we study in this paper is a collection of unordered points irregularly distributed in the 3D, which makes
the defense challenging.
%
We refer to this defense method as \textit{PointAE Defense}.


\cutsectionup
\section{Experimental Evaluation}
\cutsectiondown
In this section, we
compare our shape-aware attacks with the state-of-the-art attack methods for 3D point clouds 
against different defense mechanisms.

\cutsubsectionup
\subsection{Experimental Setup}
\cutsubsectiondown

\cutparagraphup
\paragraph{Datasets and 3D models.}
\cutparagraphdown
We conduct our experiments on the 3D point cloud classification benchmark  \textit{aligned ModelNet40}~\cite{sedaghat2017orientation, wu20153d}, and follow the same experimental settings as reported in
\cite{xiang2019generating}.
This dataset consists of $12,311$ CAD models from $40$ common object categories,
where $9,843$ objects are used for training and $2,468$ are for testing.\footnote{The compared methods and ours do not require validation, as they are optimization-based.}
We uniformly sample $1,024$ points on the surface of each object in the point space in the Cartesian coordinate (XYZ), 
and normalize them into a unit ball by centering and scaling.
To evaluate the performance of adversarial attacks, we follow the protocol in \cite{xiang2019generating} for targeted attack,
 which uses $2,250$ victim-target pairs sampled from 10 major categories 
 (airplane, bed, bookshelf, bottle, chair, monitor, sofa, table, toilet, and vase).
For untargeted attacks, we use all the $2,468$ testing examples.

The victim model for adversarial attack is the popular PointNet~\cite{qi2017pointnet}, which has
shown the state-of-the-art performance in many tasks.
For the attack transferability analysis, we also use PointNet++~\cite{qi2017pointnet++} 
and DGCNN~\cite{wang2018dynamic} as additional 3D point cloud classification models.
We train the 3D point cloud classification models
by minimizing the standard cross-entropy loss on aligned ModelNet40.

To learn the mapping from raw 3D point cloud data to the latent space, 
we leverage the point cloud auto-encoder (PointAE) with a latent space with 128 dimension approximating the shape manifold.
Here, we consider two different architectures of PointAE proposed in different literature:
MLP~\cite{achlioptas2018learning} and AtlasNet~\cite{groueix2018atlasnet}.
We pre-train PointAE on
ShapeNetCore~\cite{chang2015shapenet};
because the distribution of categories in ModelNet40 is highly imbalanced, training only on ModelNet40 
leads to a severe performance downgrade.
We note that PointAE-MLP was also trained on ShapeNetCore in \cite{achlioptas2018learning}.
In summary, we pre-train the PointAE on ShapeNetCore~\cite{chang2015shapenet} using more than $51,300$ object instances 
and fine-tune the auto-encoder on the aligned ModelNet40 dataset for $700$ epochs.
We find that such shape pre-training is crucial to learn a good latent space with a low reconstruction error, 
and effective shape deformation and interpolation in the latent space.

\begin{table}[t]
	\centering
	\scalebox{0.9}{
	\begin{tabular}{l|c|c|c|c}
		\toprule
		Classifier / Defense &  Clean & SOR~\cite{zhou2019dup} & PointAE-MLP & PointAE-AtlasNet \cr
		\midrule
		PointNet & 88.7\% & 88.5\% & 86.8\% & 87.4\% \cr
		PointNet++ & 92.7\% & 92.4\% & 85.0\%  & 86.4\% \cr 
		DGCNN & 91.0\% & 90.8\% & 83.7\% & 85.6\% \cr
		Different initialization & 90.3\% & 90.0\% & 86.0\% & 86.9\%  \cr
		\bottomrule
	\end{tabular}
	}
	\cuttablecaptionup
	\caption{Classification accuracy on ModelNet40 with PointAE Defense in Eq.~\eqref{eqn:ae}.
	}
	\cuttabledown
	\cuttabledown
	\label{tab:exp-benchmark-cls-acc}
\end{table}

\cutparagraphup
\paragraph{Attack and defense methods.}
\cutparagraphdown
We compare the state-of-the-art adversarial attacks for 3D point clouds with our proposed methods.
Point shifting attack (Shift-Point)~\cite{xiang2019generating} perturbs the victim point cloud 
on the point space while minimizing the perturbation magnitude under the $\ell_2$ constraint.
Point addition attacks~\cite{xiang2019generating} introduce new points to the original victim point cloud, 
where we compare three strategies:
(1) adding a number of points independently (Add-Point), 
(2) adding several clusters of points (Add-Cluster), and
(3) adding several point clouds from another object belonging to a different category (Add-Object).

As described in Section~\ref{sec:method}, we introduce three variants of the shape-aware adversarial attacks 
by injecting adversarial perturbations in the latent space.
We additionally constrain the perturbation by
(1) minimizing the $\ell_2$ distance in the latent space (ShapeAdv-Latent~$\ell_2$),
(2) minimizing the Chamfer distance in the point space (ShapeAdv-Chamfer), and
(3) minimizing the Chamfer distance from the auxiliary point clouds sampled 
from the category of the input as well as the input (ShapeAdv-Auxiliary).

For the defense experiments, we consider the sparse outlier removal (SOR)~\cite{rusu2008towards,zhou2019dup} 
and our proposed auto-encoder based method.
SOR is a simple defense method that removes outlier points in the point space, which is shown to be effective in \cite{zhou2019dup};
for each point, it first computes the average distance from its $k$-nearest neighbors, and then filters out points 
if the average distance is larger than $\mu + c \cdot \sigma$, where $\mu$ and $\sigma$ are the mean 
and standard deviation of the average distances, and $k$ and $c$ are hyperparameters.
%
We set the hyperparameters to be $k=2$ and $c=2.5$, which is more conservative than the original paper~\cite{zhou2019dup} 
and shows a strong defense performance against the adversarial point clouds.
%
In PointAE Defense, as described in Section~\ref{sec:defense}, each point cloud in the test dataset is auto-encoded 
and then considered as an input of the classification model, as in Eq.~\eqref{eqn:ae}.
Specifically, for each point cloud in the test dataset, we use the reconstruction-based method described in Section~\ref{sec:defense}:
using the direct output from the encoder model for decoding as in Eq.~\eqref{eqn:ae}, referred as PointAE Defense.
%
To show the classification performance after applying defense mechanisms, 
we report the test accuracy with the reconstructed point clouds in Table~\ref{tab:exp-benchmark-cls-acc}.

\cutparagraphup
\paragraph{Evaluation metrics.}
\cutparagraphdown
We use both attack success rate as well as the perturbation magnitude measure (using the Chamfer distance) as our major evaluation metrics.
For attack success rate, we compute the number of successfully attacked object instances divided by the total number.
For the Chamfer distance measure, we report the quantitative results in the \textit{best}, \textit{average}, and \textit{worst} performance over category-wise performances,
as some category pairs are more difficult to attack than others.
%
The \textit{best case} represents the most easily attacked victim class, \textit{average case} for attacking all classes, and \textit{worst case} for the most difficult class.

\begin{table}[t]
	\centering
	\cuttableup
	\scalebox{0.85}{
	\begin{tabular}{l|c|c|c|c}
	    \toprule
		Attack Type & \multicolumn{2}{c|}{Targeted} & \multicolumn{2}{c}{Untargeted} \cr
        \hline
		Attack / Decoder  & MLP & AtlasNet & MLP & AtlasNet \cr
        \hline
		Latent~$\ell_2$ (Eq.~\eqref{eqn:shapeadv_l2})
		& 2.99 / 8.56 / 14.75 & 1.97 / 4.29 / 6.59 & 2.99 / 6.26 / 11.12 & 1.44 / 3.26 / 5.52\cr
		\hline
		Chamfer (Eq.~\eqref{eqn:shapeadv_ch})
		& 2.14 / 4.55 / 8.82 & 1.46 / 3.15 / 5.15 & 1.36 / 3.47 / 5.68 & 1.00 / 2.70 / 4.42\cr
		\hline
		Auxiliary (Eq.~\eqref{eqn:shapeadv_aux})
		&
		2.47 / 5.16 / 9.61 & 2.25 / 4.12 / 6.10 &
		1.46 / 4.61 / 8.27 & 1.20 / 3.86 / 6.46\cr
		\bottomrule
	\end{tabular}
	}
	\cuttablecaptionup
	\caption{
		Quantitative evaluation of ShapeAdv under \textbf{targeted} attacks.
		We report the perturbation magnitude in the Chamfer distance ($\times 10^{-3}$) (the lower the better).
		%
		We report three different metrics: \textit{best}, \textit{average} and \textit{worst}.
		We note that the attack success rate of all methods is 100\%.
	}
	\cuttabledown
	\cuttabledown
	\label{tab:exp-shapeadv-all}
\end{table}

\begin{figure}[t]
\centering
\scriptsize\setlength{\tabcolsep}{0mm}
\begin{tabular}{
>{\centering}m{14mm}
>{\centering}m{\figwidth}
>{\centering}m{\figwidth}
>{\centering}m{\figwidth}
>{\centering}m{\figwidth}
>{\centering}m{\figwidth}
>{\centering}m{\figwidth}
}
\multirow{2}{*}{ShapeAdv} &
\multicolumn{2}{c}{Airplane $\rightarrow$ Bottle} &
\multicolumn{2}{c}{Bottle $\rightarrow$ Chair} &
\multicolumn{2}{c}{Chair $\rightarrow$ Airplane} \cr
& Front & Side & Front & Top & Front & Side \cr
\hline
No Attack &
\includegraphics[width=\figwidth]{{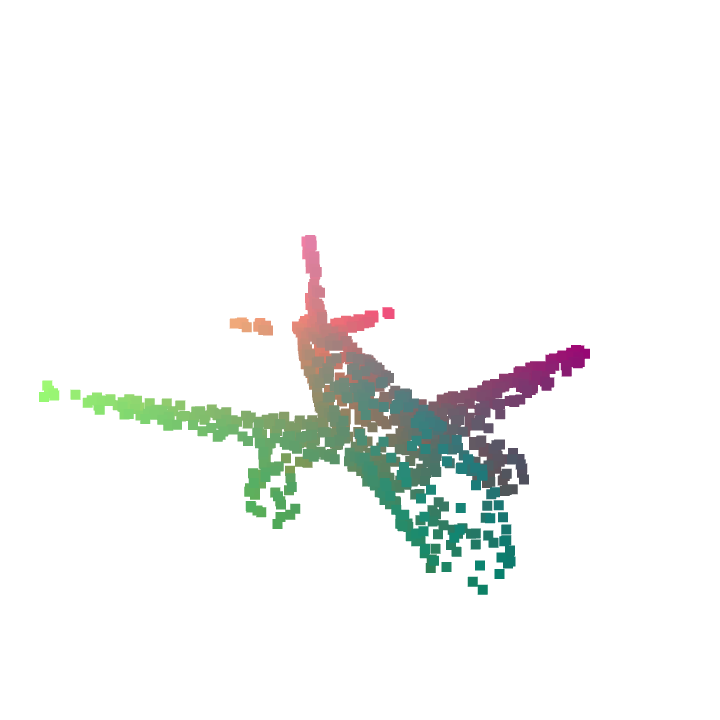}} &
\includegraphics[width=\figwidth]{{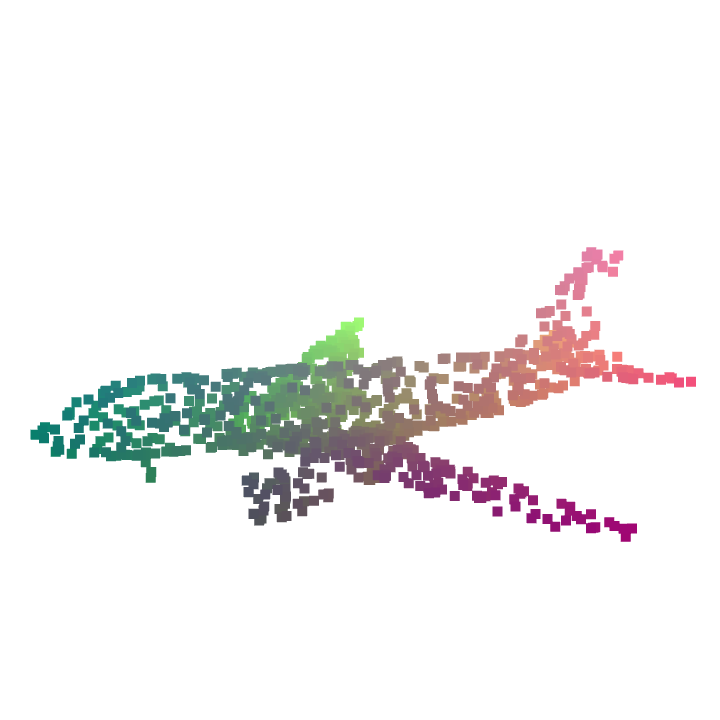}} &
\includegraphics[width=\figwidth]{{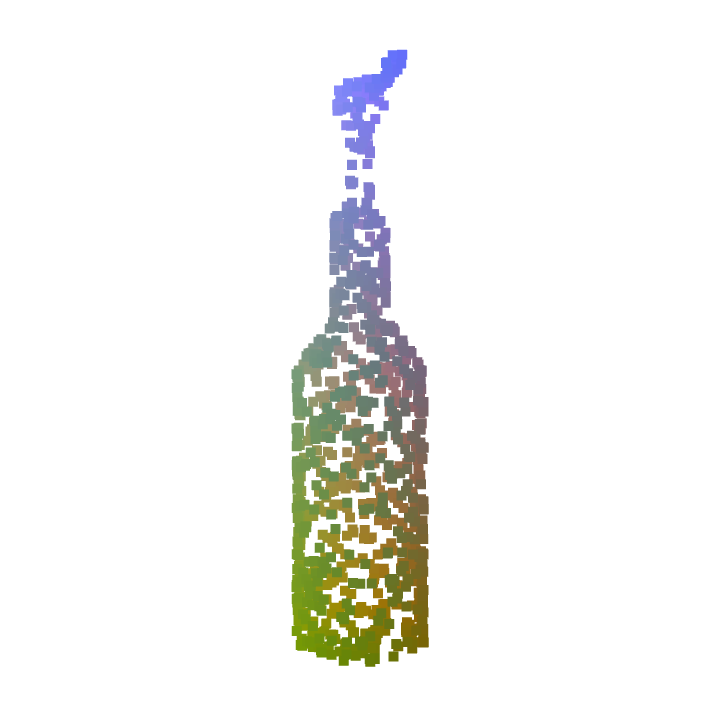}} &
\includegraphics[width=\figwidth]{{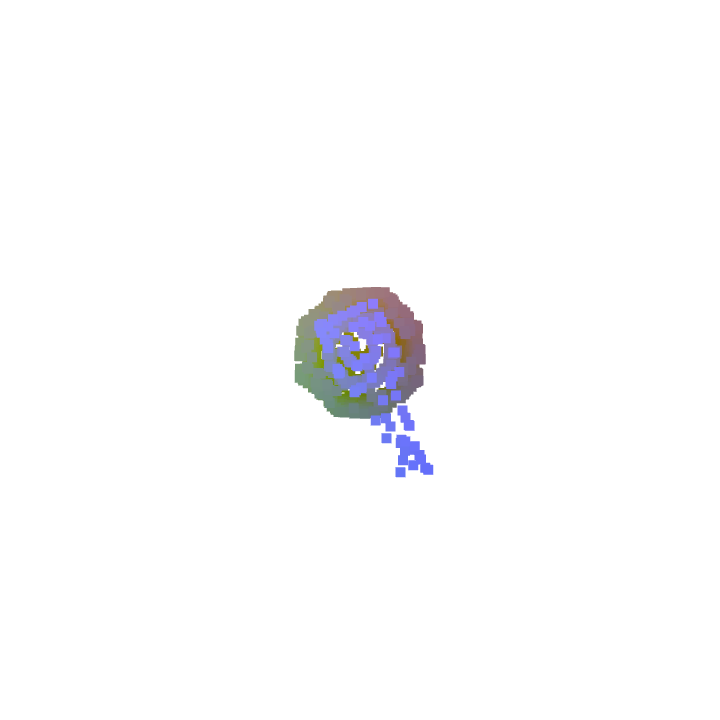}} &
\includegraphics[width=\figwidth]{{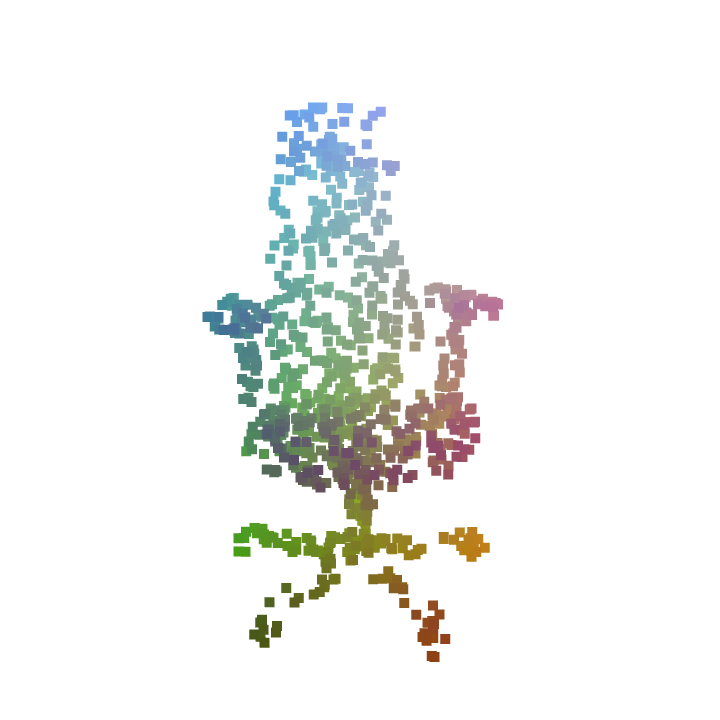}} &
\includegraphics[width=\figwidth]{{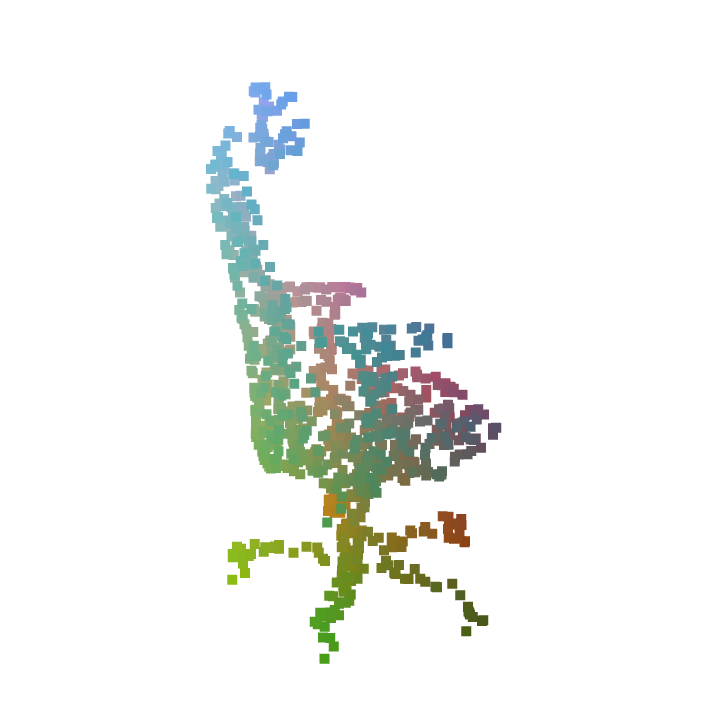}} \cr
\hline
MLP-Latent~$\ell_2$ &
\includegraphics[width=\figwidth]{{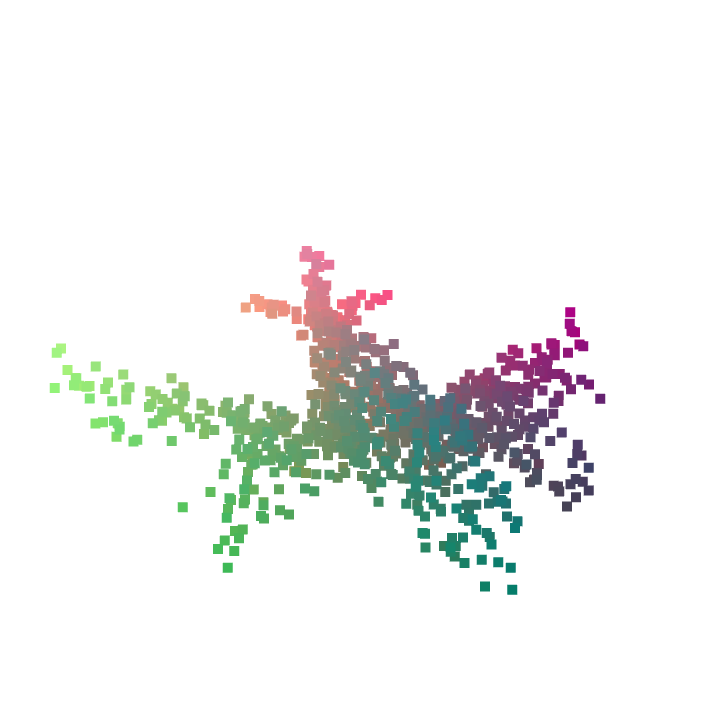}} &
\includegraphics[width=\figwidth]{{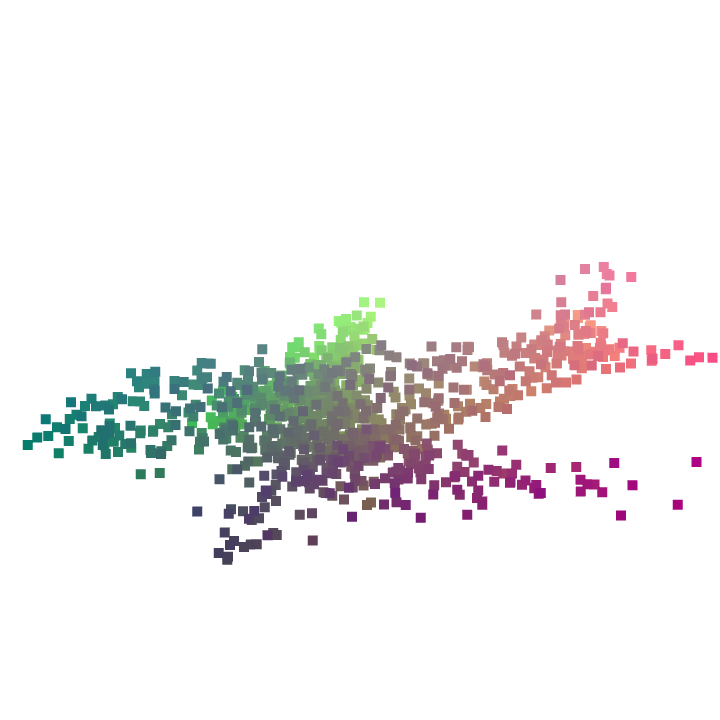}} &
\includegraphics[width=\figwidth]{{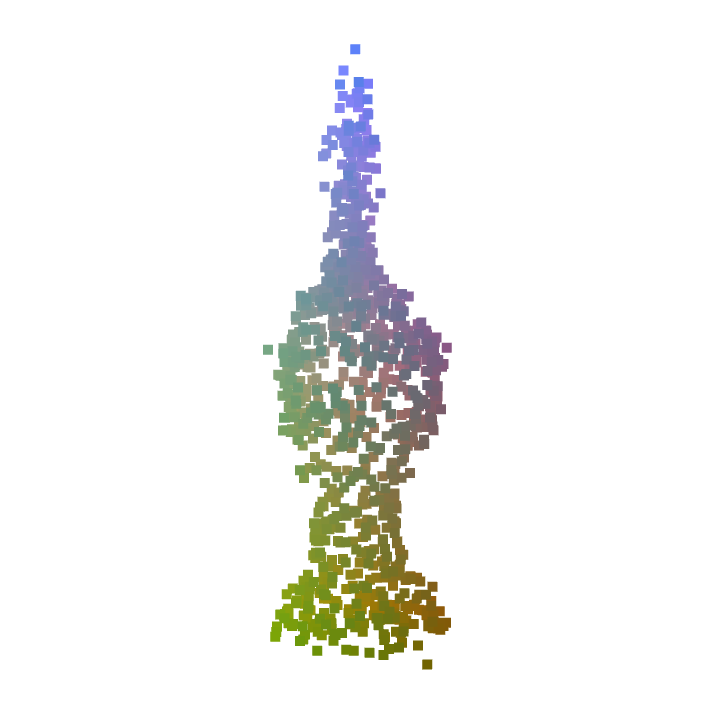}} &
\includegraphics[width=\figwidth]{{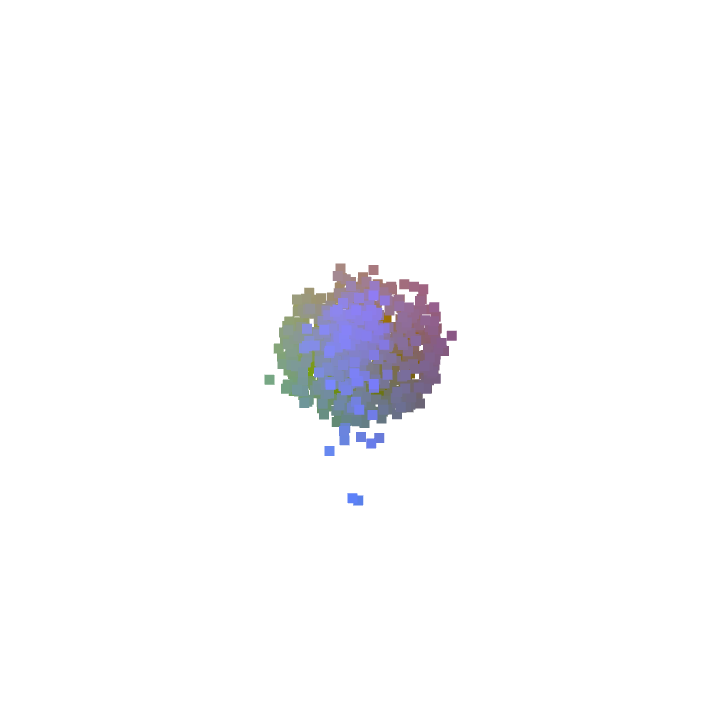}} &
\includegraphics[width=\figwidth]{{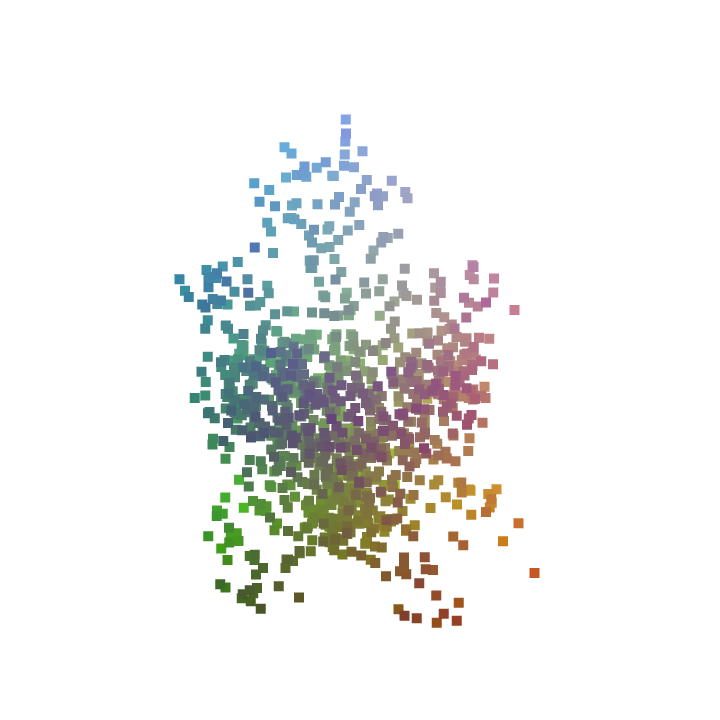}} &
\includegraphics[width=\figwidth]{{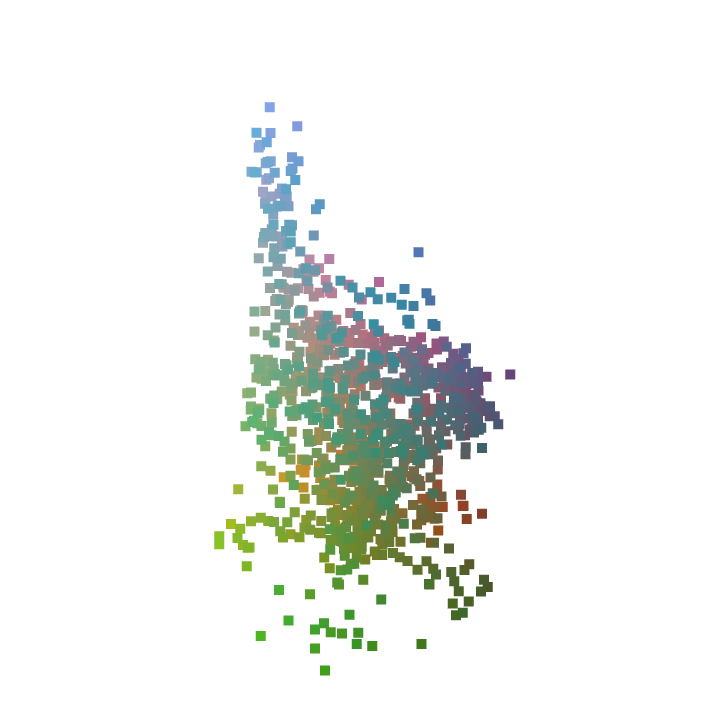}} \cr
MLP-Chamfer &
\includegraphics[width=\figwidth]{{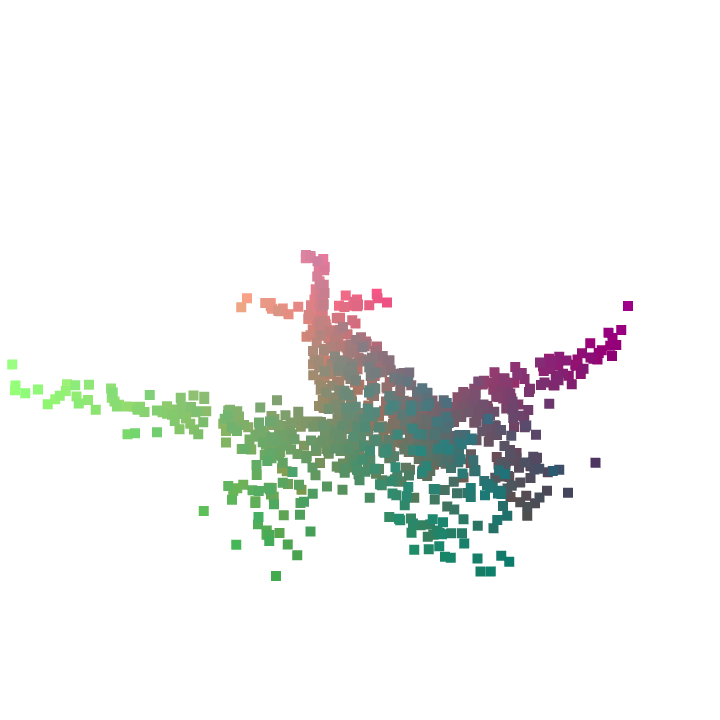}} &
\includegraphics[width=\figwidth]{{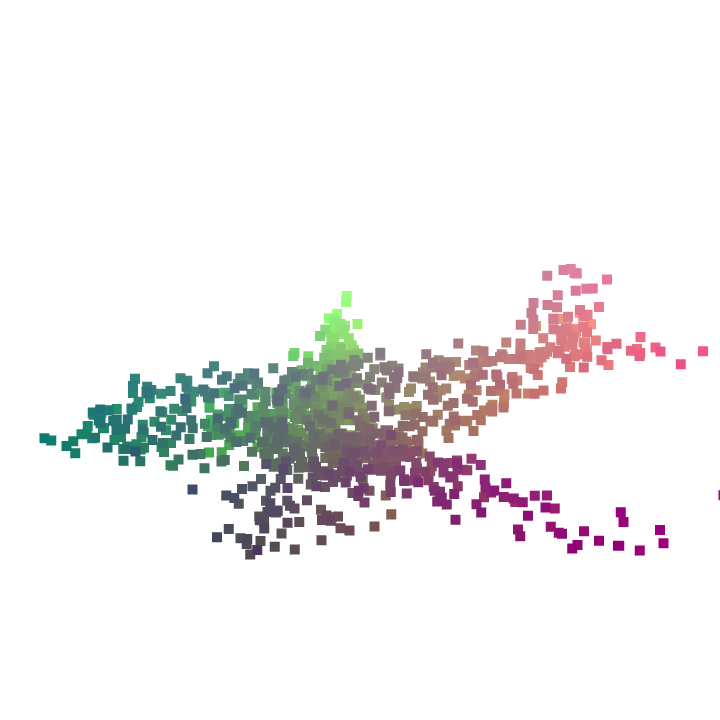}} &
\includegraphics[width=\figwidth]{{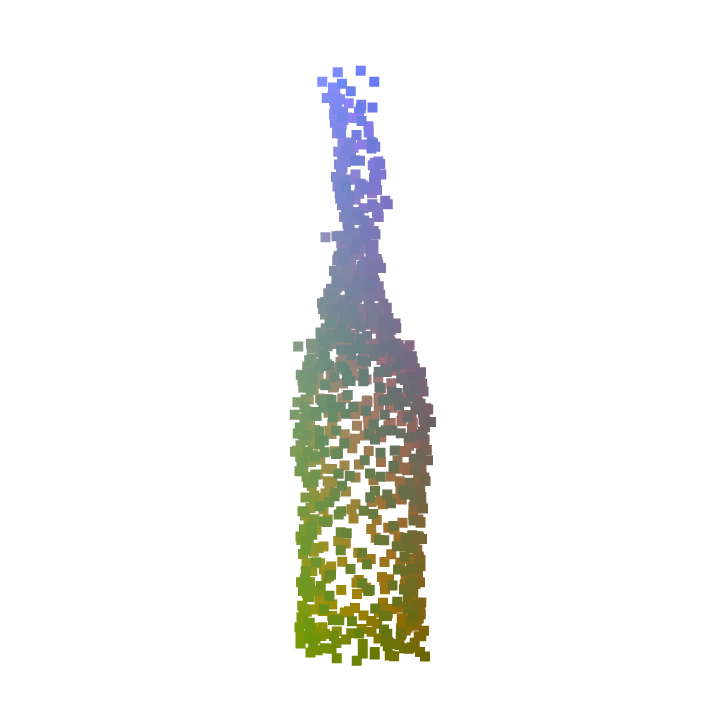}} &
\includegraphics[width=\figwidth]{{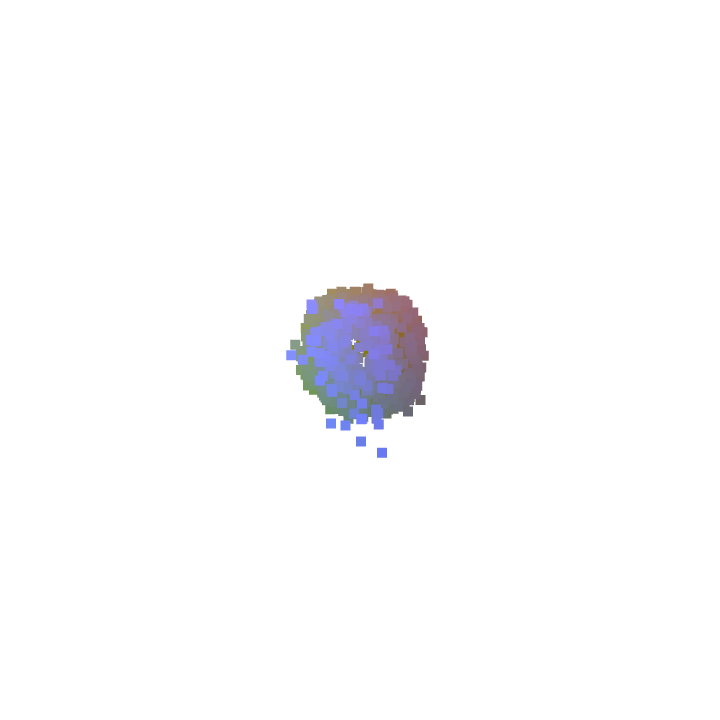}} &
\includegraphics[width=\figwidth]{{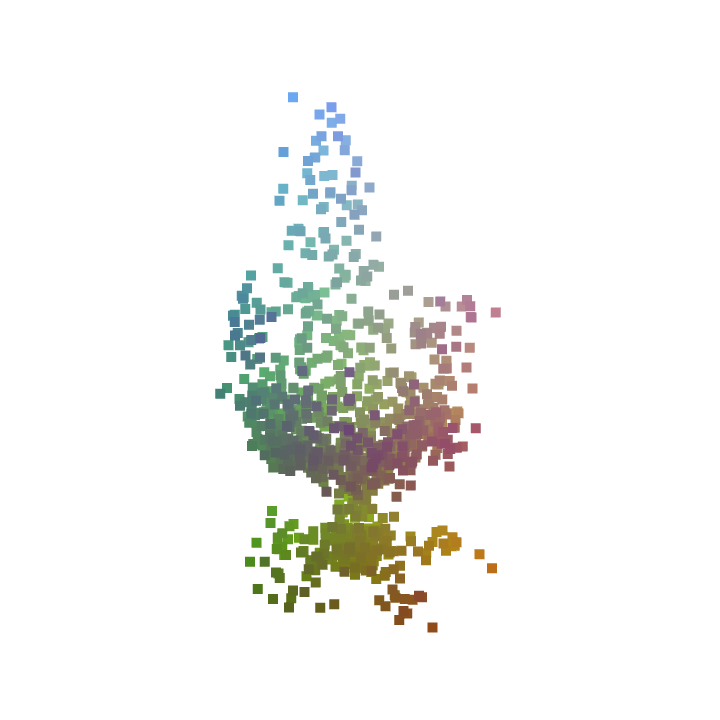}} &
\includegraphics[width=\figwidth]{{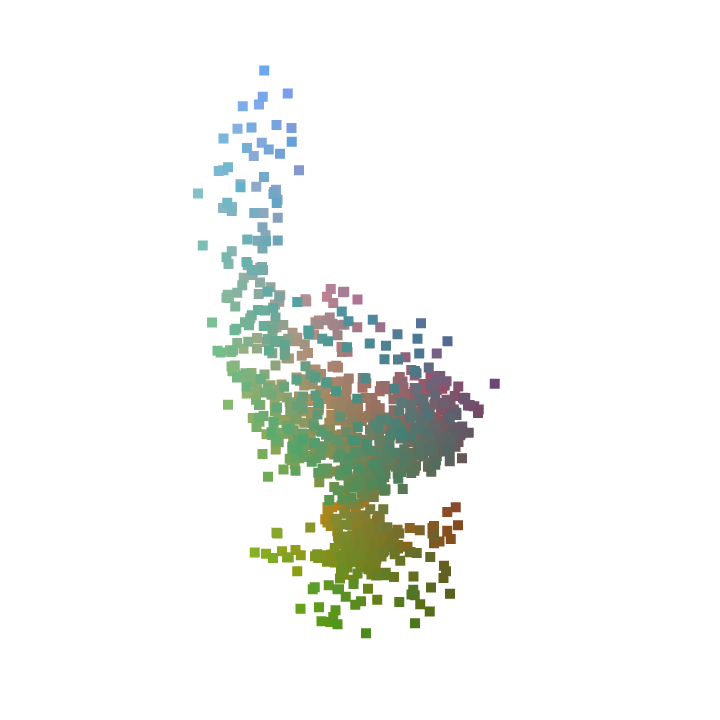}} \cr
\hline
Atlas-Latent~$\ell_2$ &
\includegraphics[width=\figwidth]{{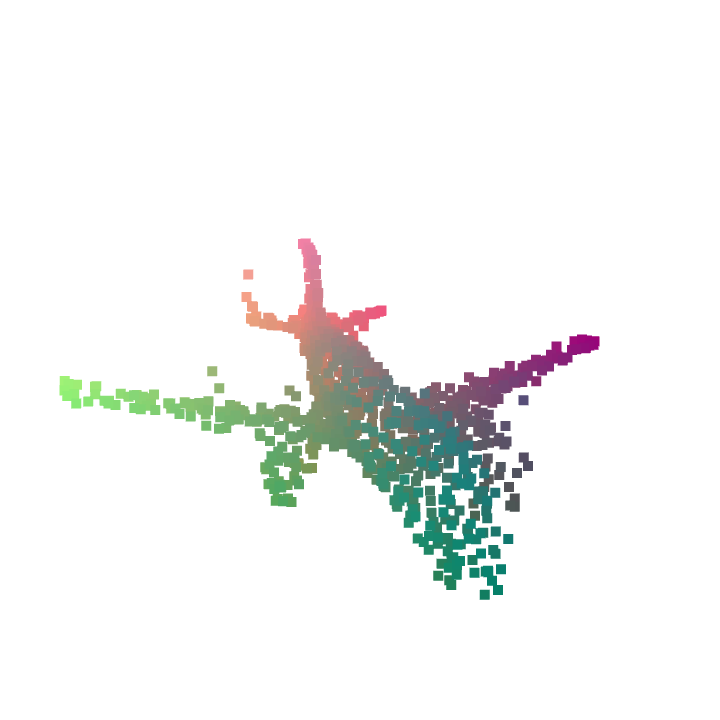}} &
\includegraphics[width=\figwidth]{{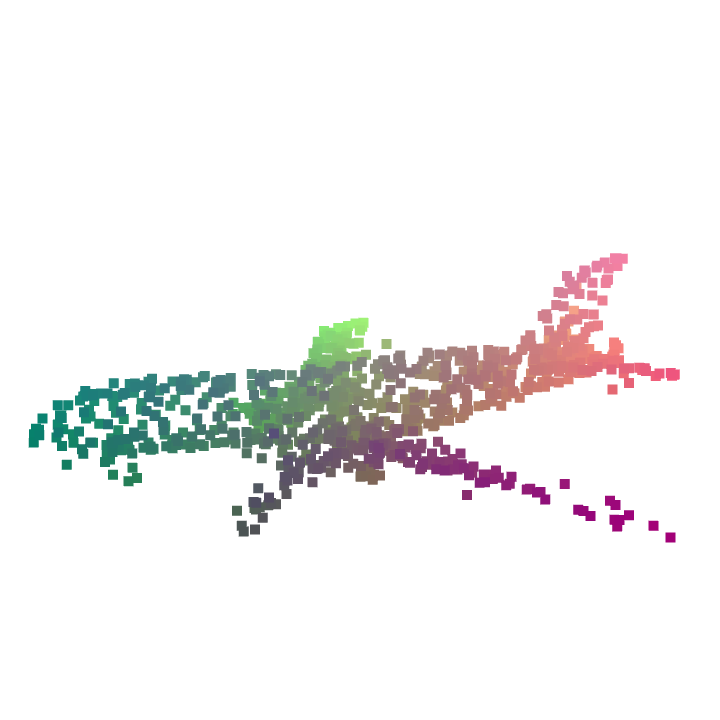}} &
\includegraphics[width=\figwidth]{{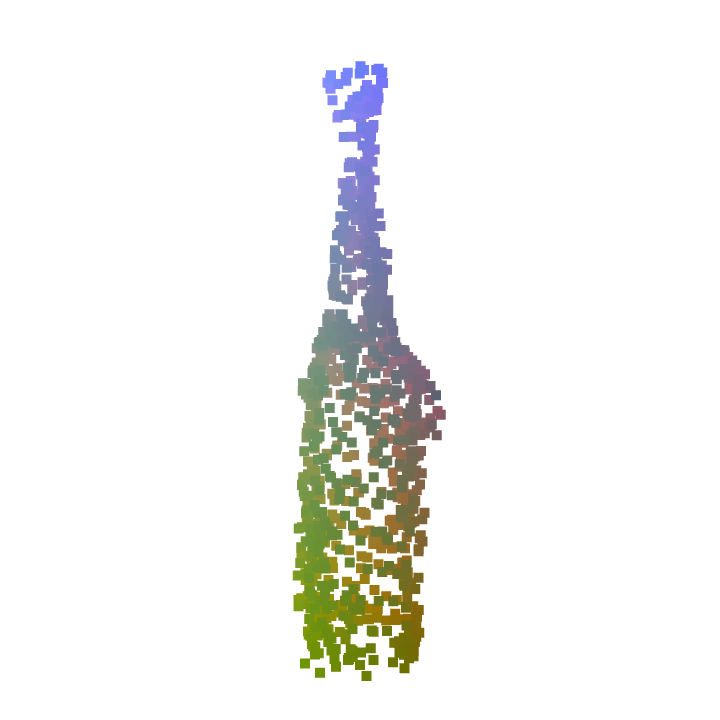}} &
\includegraphics[width=\figwidth]{{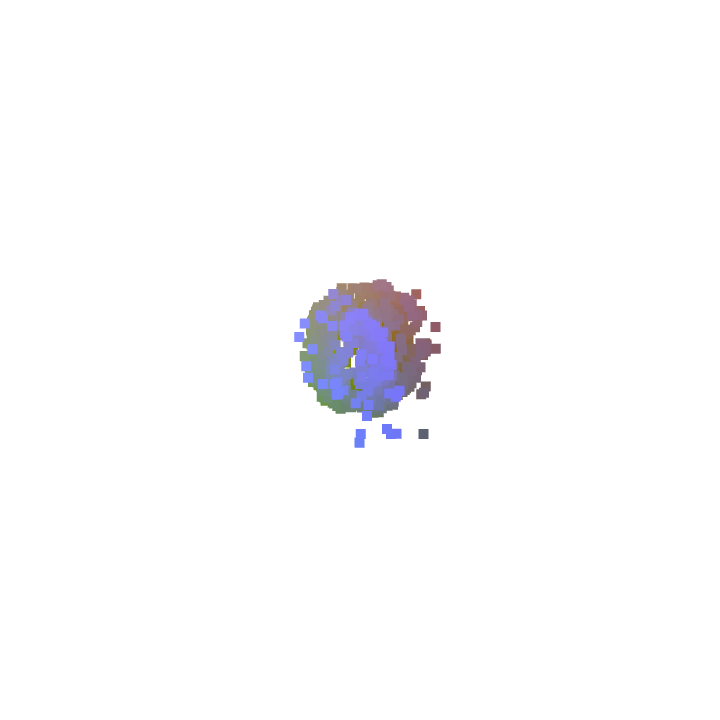}} &
\includegraphics[width=\figwidth]{{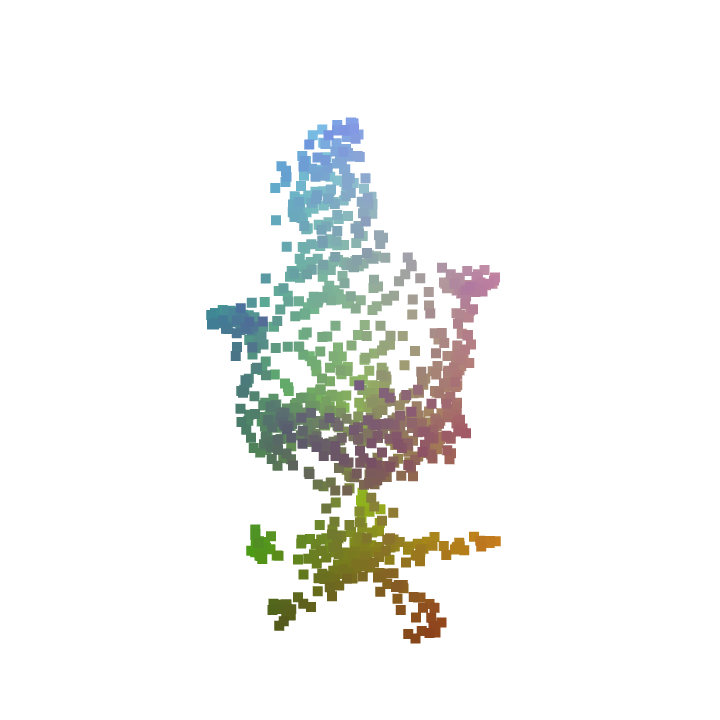}} &
\includegraphics[width=\figwidth]{{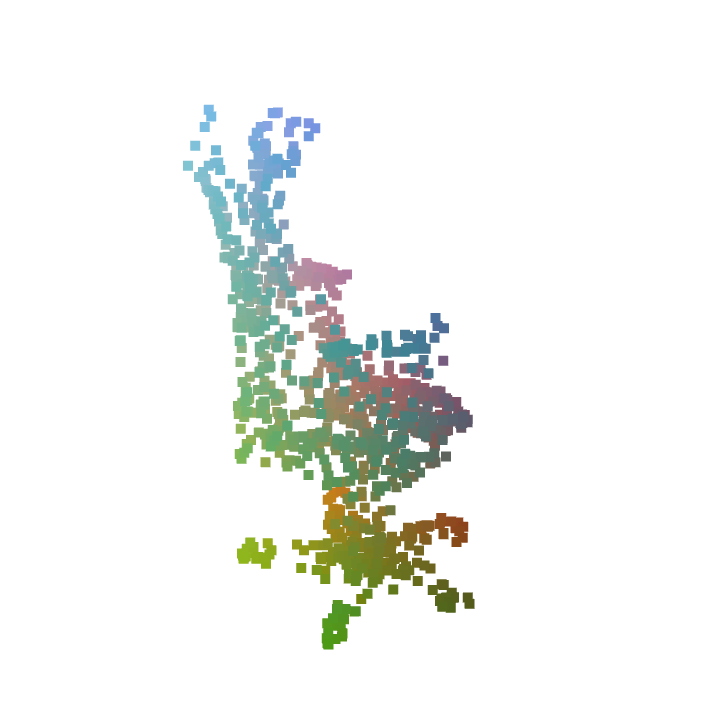}} \cr
Atlas-Chamfer &
\includegraphics[width=\figwidth]{{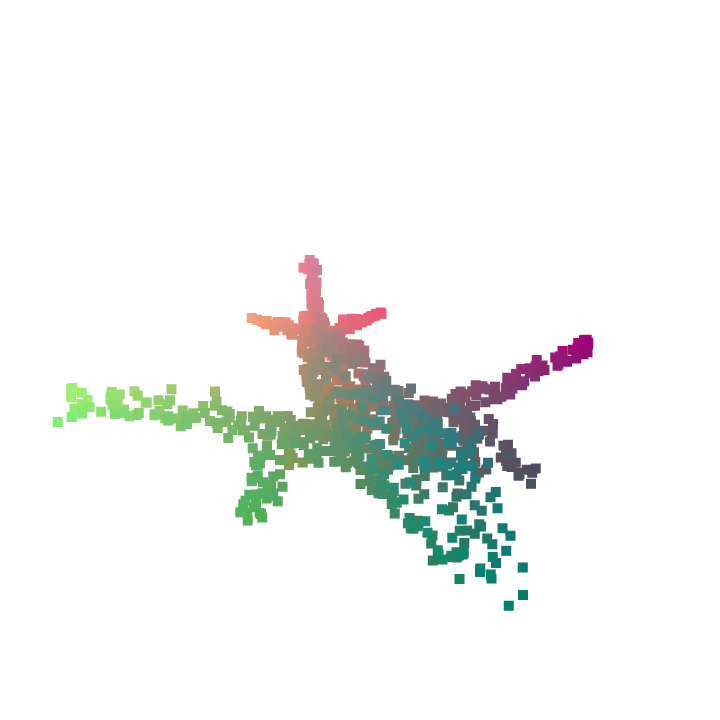}} &
\includegraphics[width=\figwidth]{{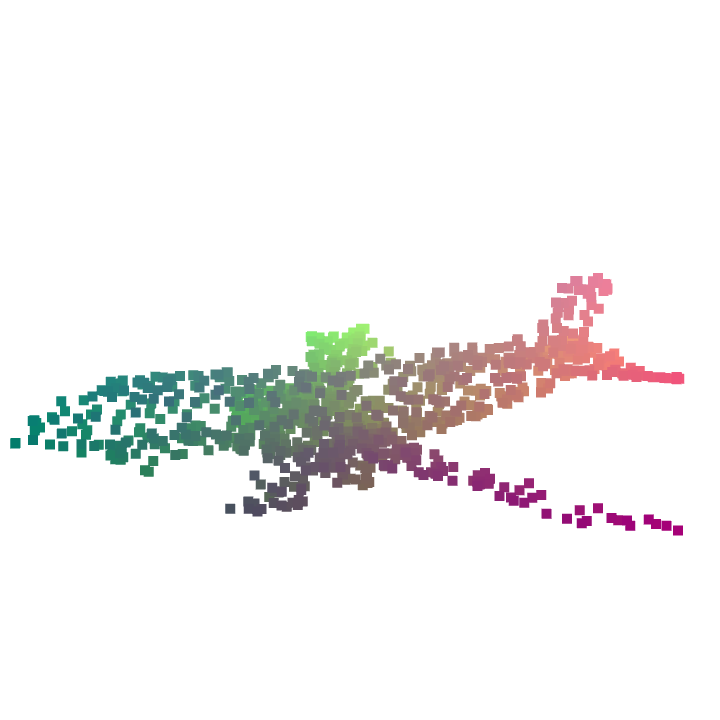}} &
\includegraphics[width=\figwidth]{{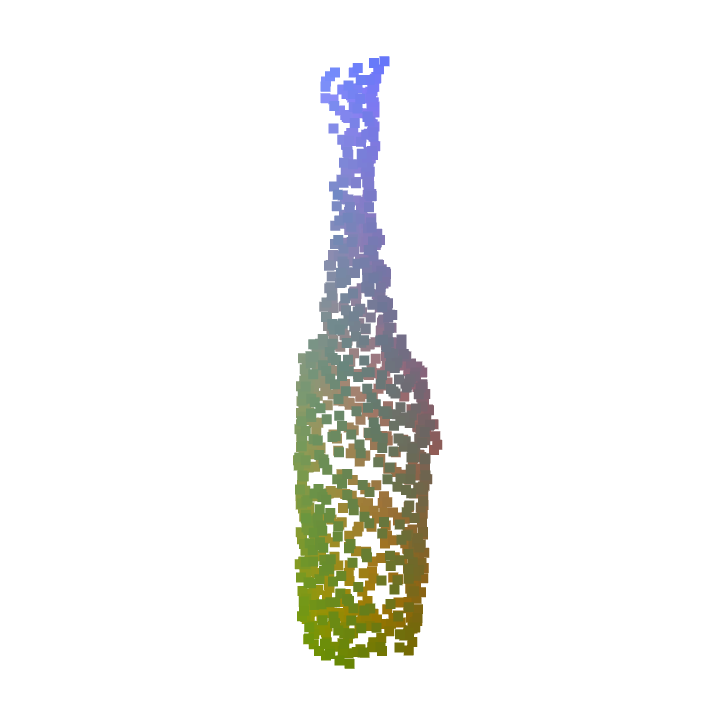}} &
\includegraphics[width=\figwidth]{{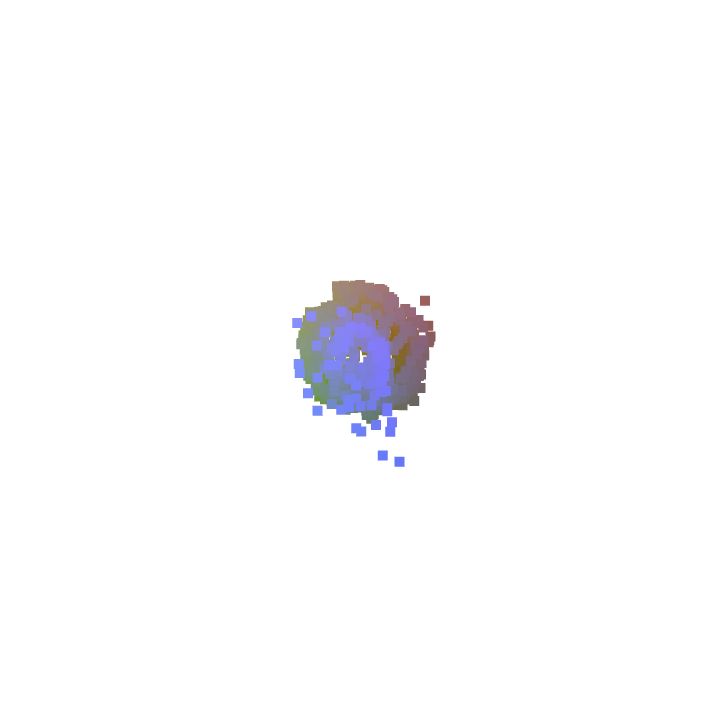}} &
\includegraphics[width=\figwidth]{{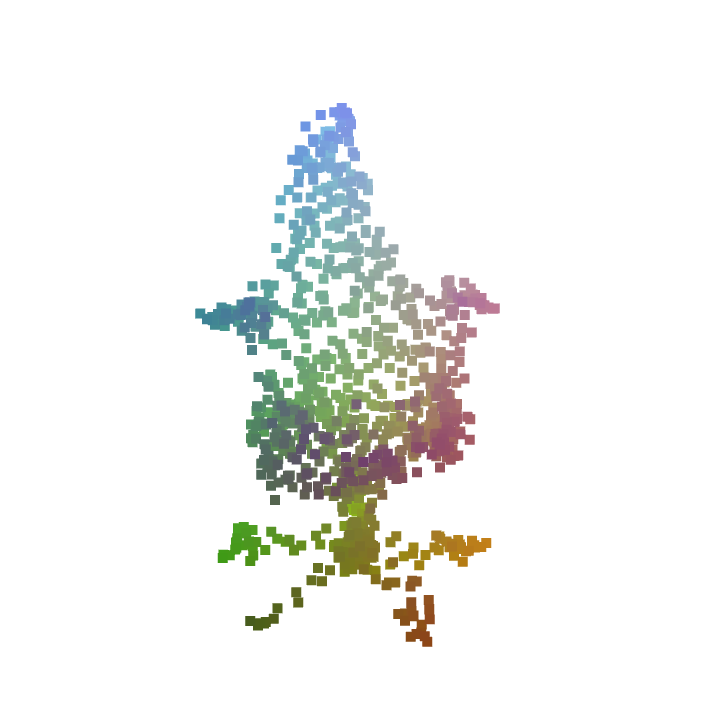}} &
\includegraphics[width=\figwidth]{{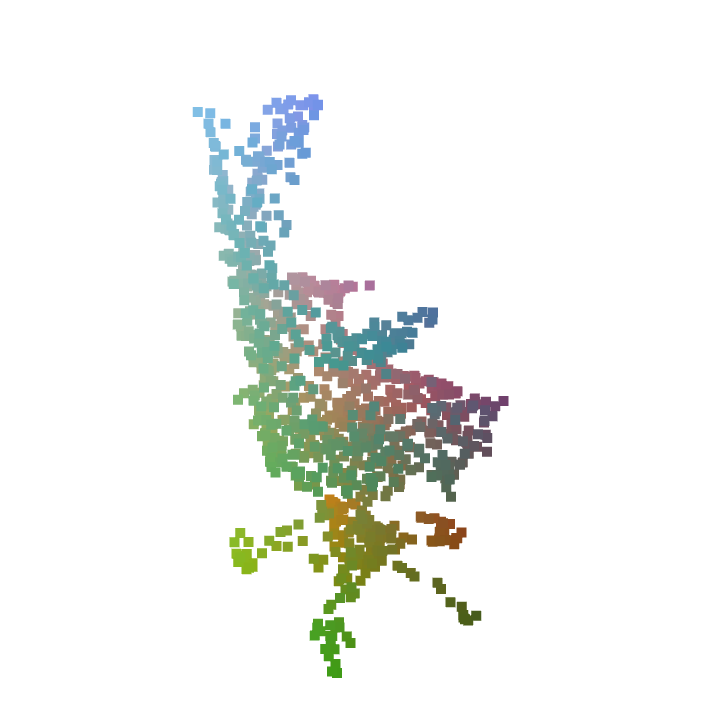}} \cr
\end{tabular}
\cutcaptionup
\caption{Qualitative comparison of ShapeAdv with PointAE-MLP and PointAE-AtlasNet under \textbf{targeted} attacks. As both models generate perceptually plausible point clouds, the adversaries of PointAE-AtlasNet preserve more details.
%
%
%
%
}
\cutcaptiondown
\label{fig:autoencoder}
\end{figure}
\begin{figure}[!t]
\centering
\scriptsize\setlength{\tabcolsep}{0mm}
\begin{tabular}{
>{\centering}m{14mm}
>{\centering}m{\figwidth}
>{\centering}m{\figwidth}
>{\centering}m{\figwidth}
>{\centering}m{\figwidth}
>{\centering}m{\figwidth}
>{\centering}m{\figwidth}
}
\multirow{2}{*}{\parbox{14mm}{\centering ShapeAdv Auxiliary}} &
\multirow{2}{*}{Bottle} &
\multirow{2}{*}{Chair} &
\multirow{2}{*}{Monitor} &
\multirow{2}{*}{Sofa} &
\multirow{2}{*}{Table} &
\multirow{2}{*}{Vase} \cr\cr
\hline
Input &
\includegraphics[width=\figwidth]{{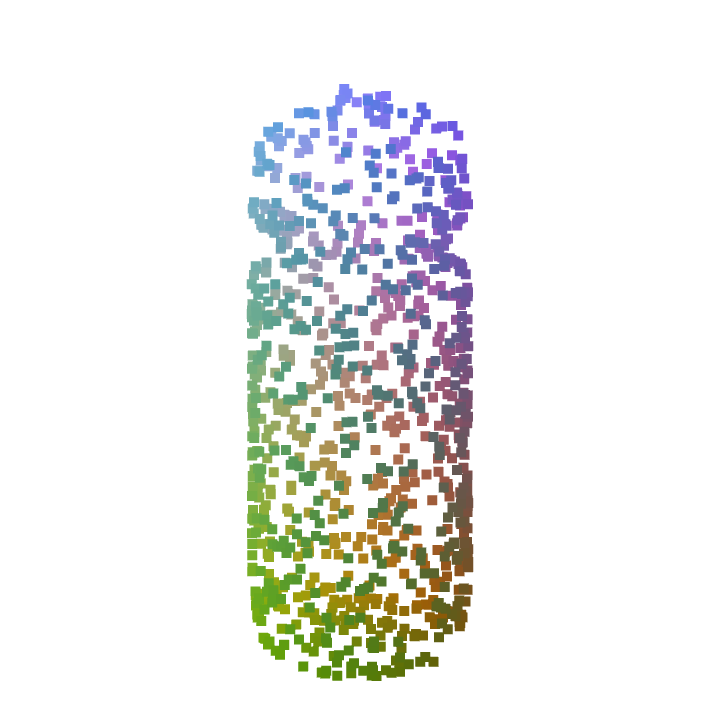}} &
\includegraphics[width=\figwidth]{{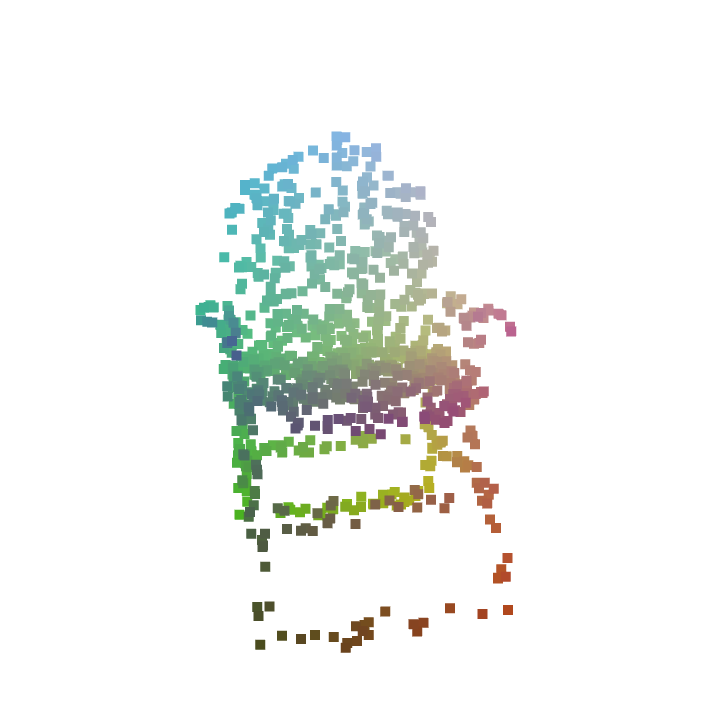}} &
\includegraphics[width=\figwidth]{{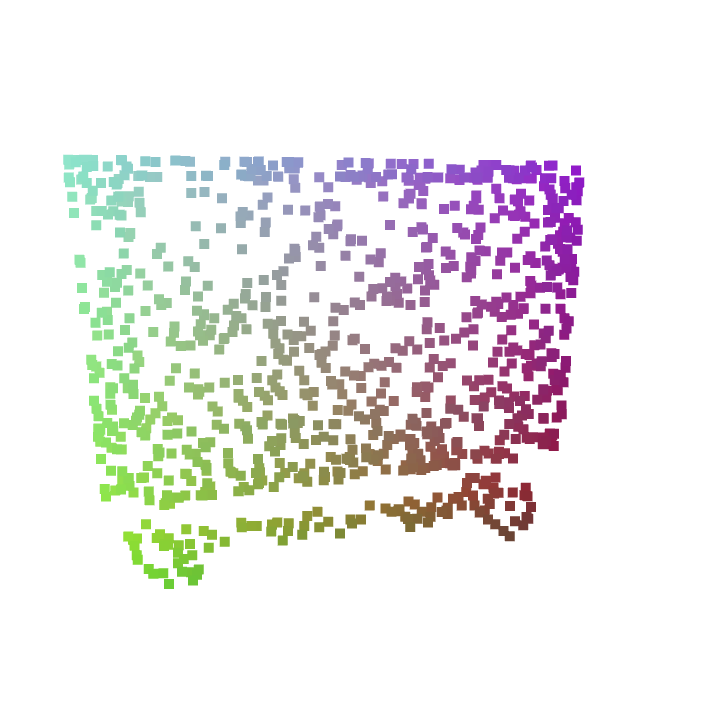}} &
\includegraphics[width=\figwidth]{{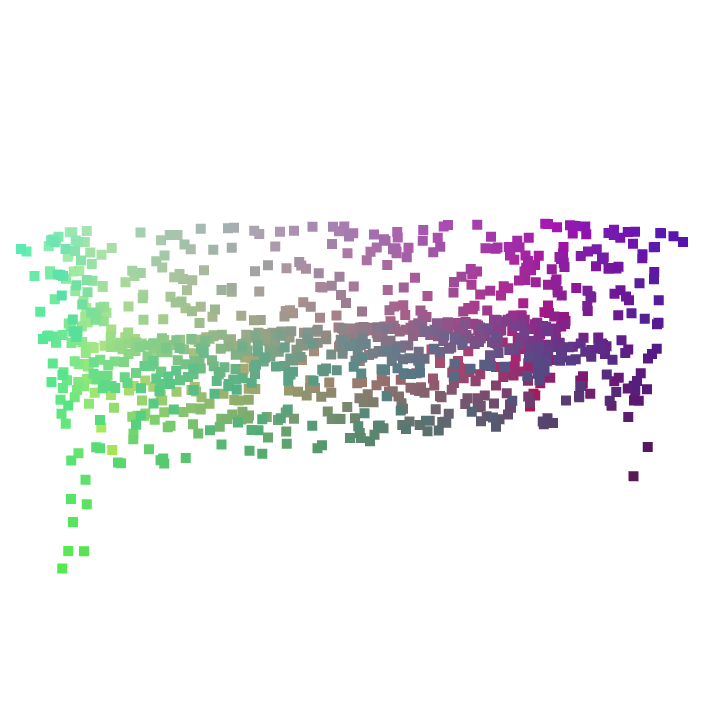}} &
\includegraphics[width=\figwidth]{{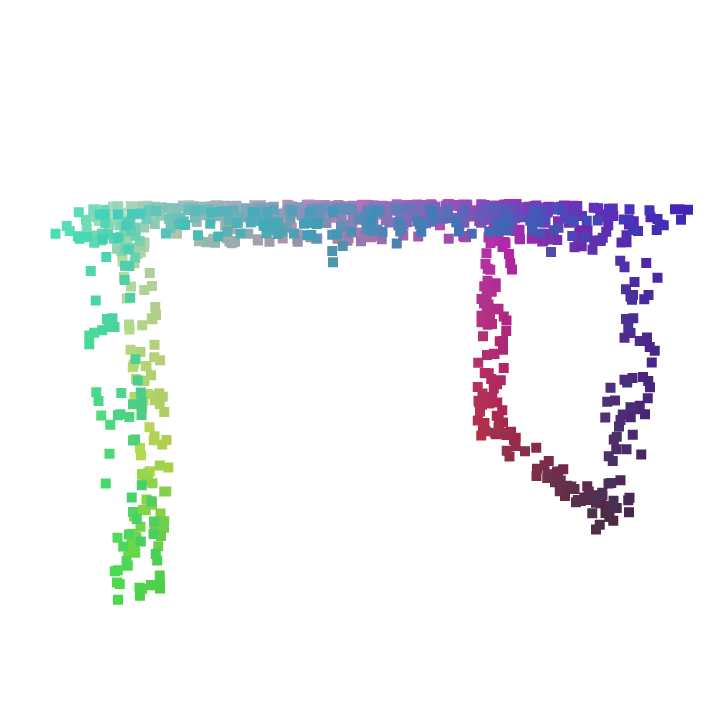}} &
\includegraphics[width=\figwidth]{{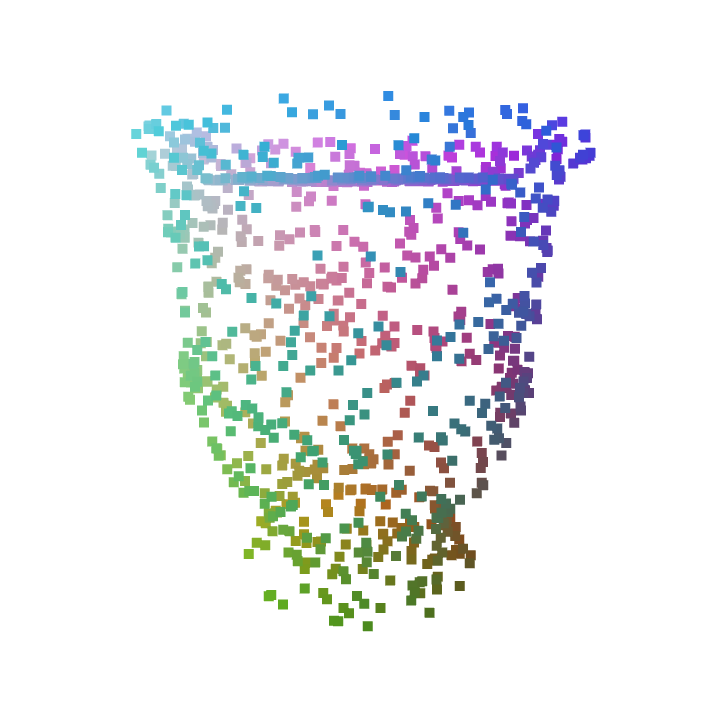}} \cr
Auxiliary &
\includegraphics[width=\figwidth]{{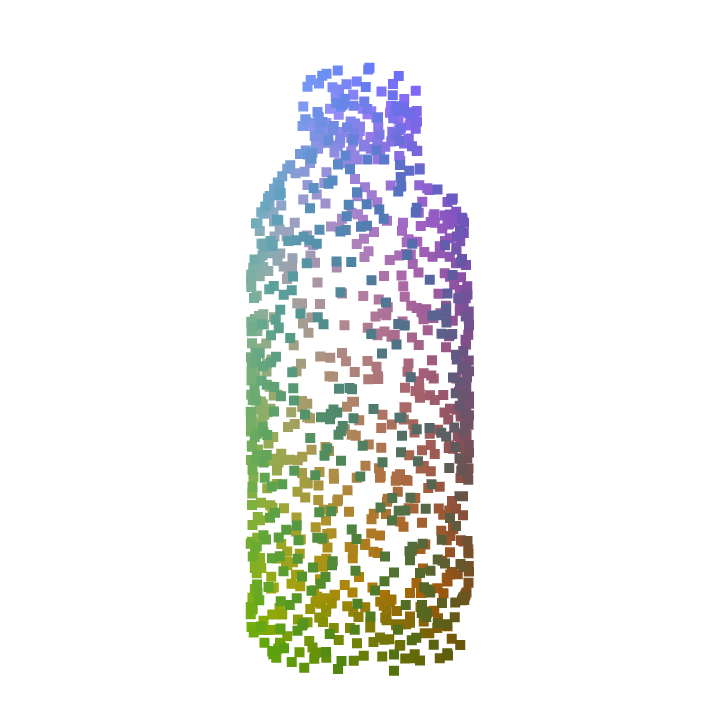}} &
\includegraphics[width=\figwidth]{{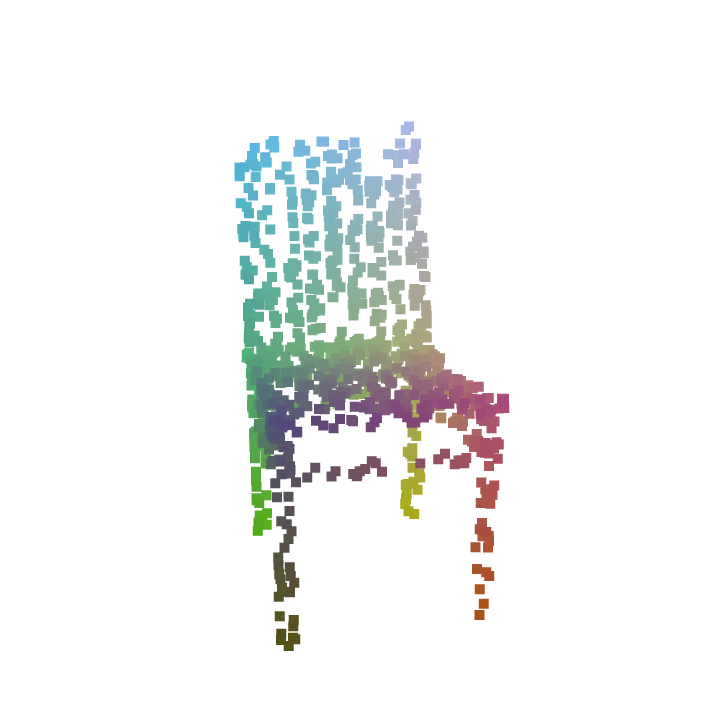}} &
\includegraphics[width=\figwidth]{{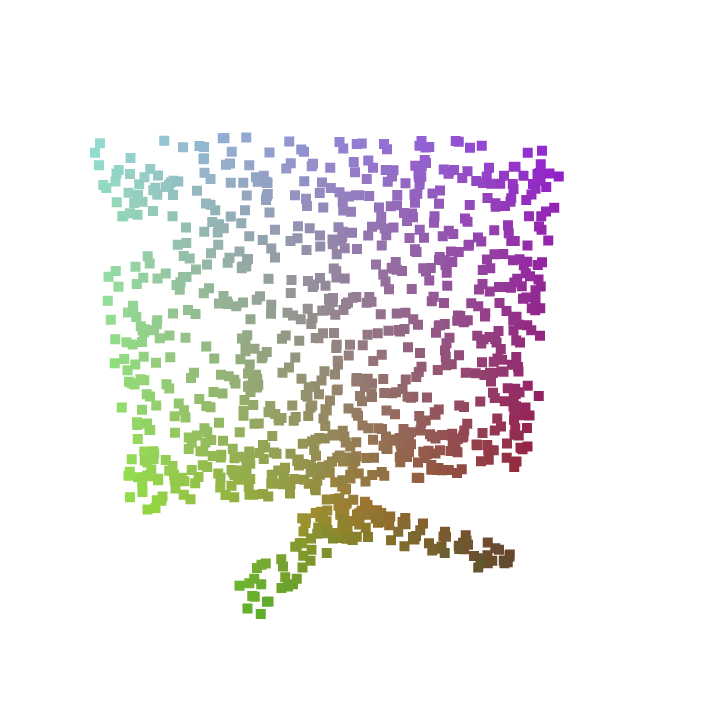}} &
\includegraphics[width=\figwidth]{{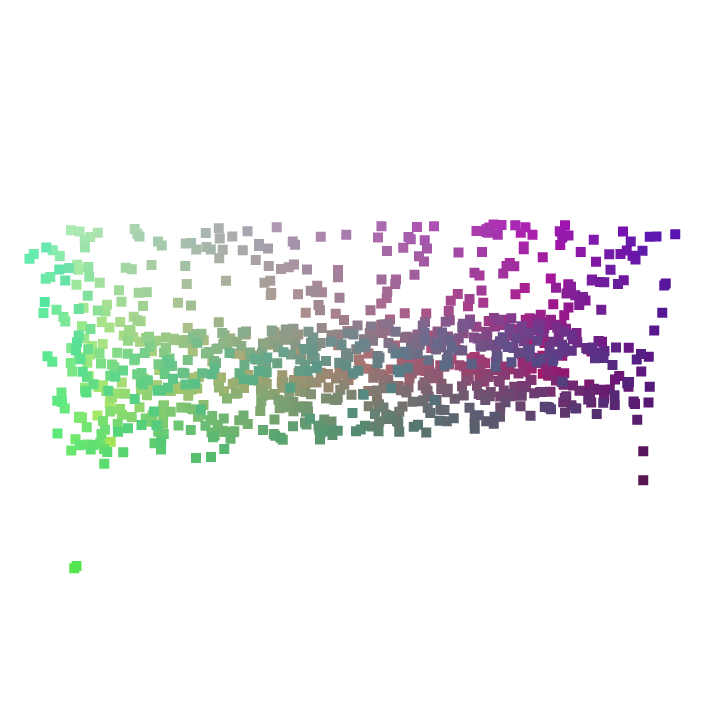}} &
\includegraphics[width=\figwidth]{{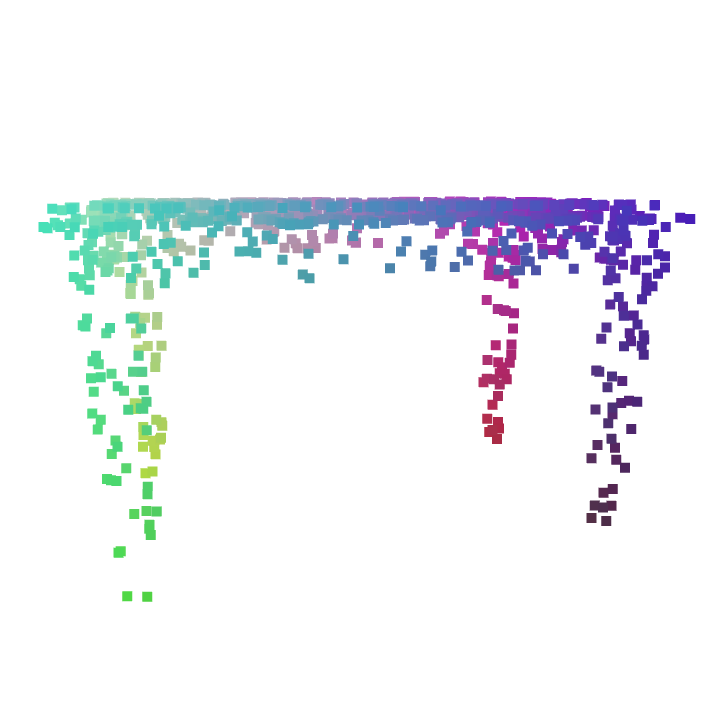}} &
\includegraphics[width=\figwidth]{{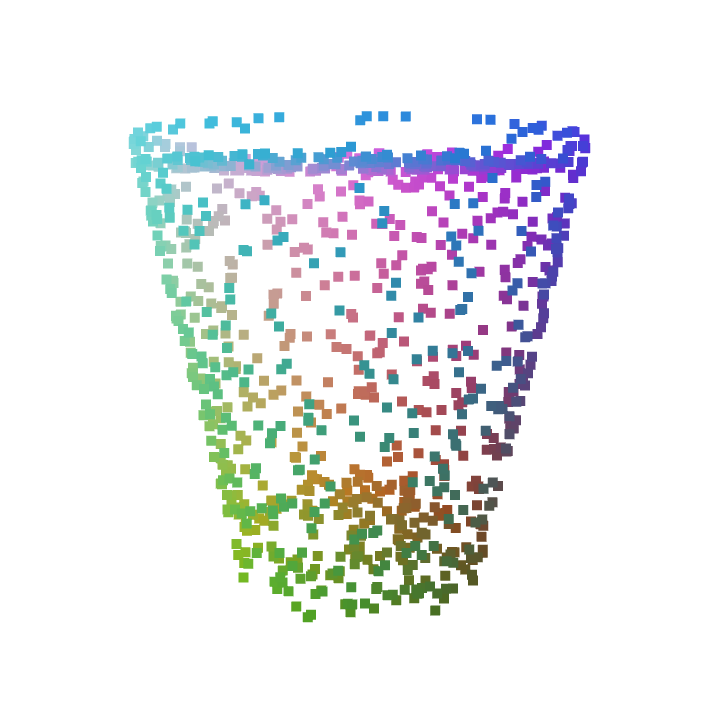}} \cr
\hline
PointAE-MLP &
\includegraphics[width=\figwidth]{{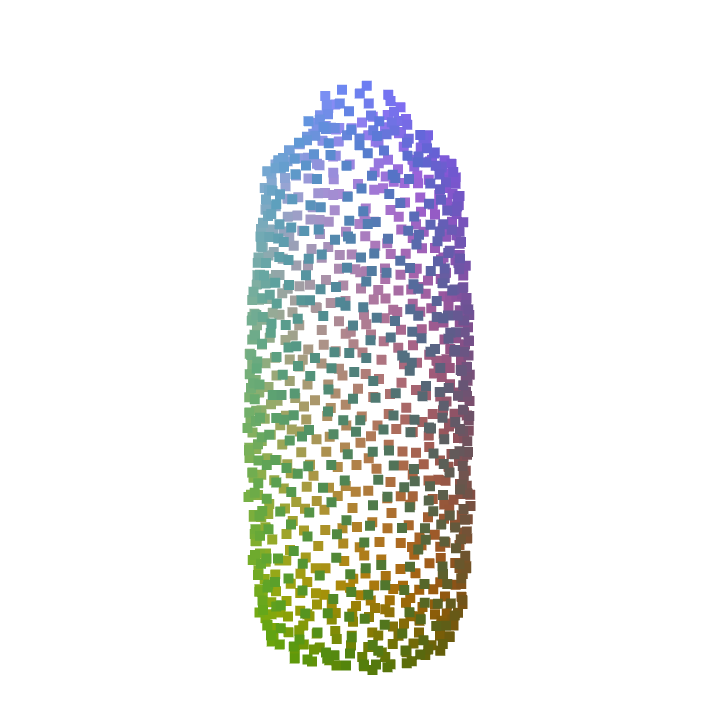}} &
\includegraphics[width=\figwidth]{{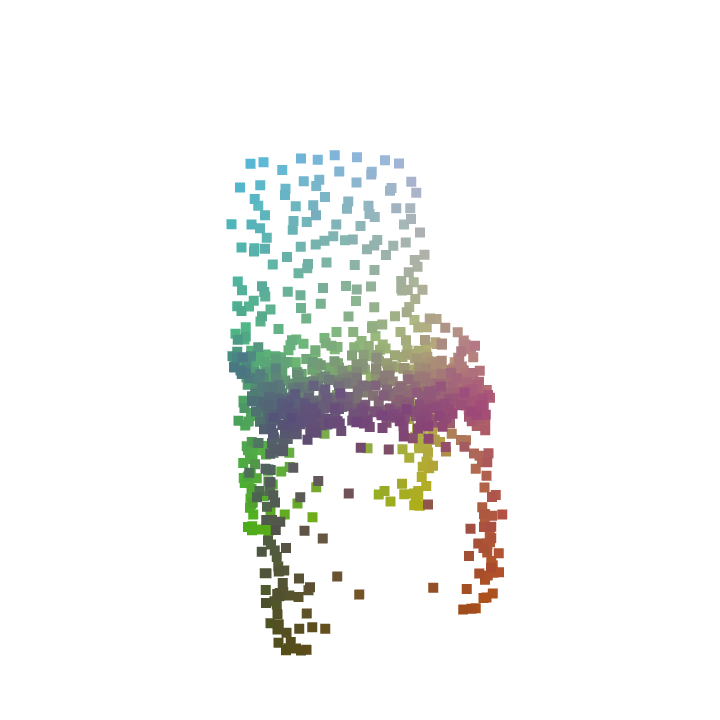}} &
\includegraphics[width=\figwidth]{{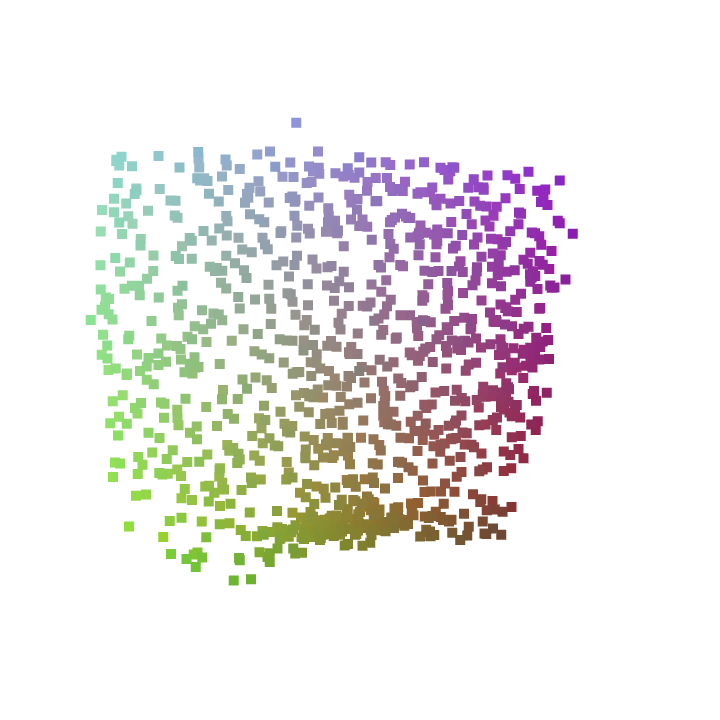}} &
\includegraphics[width=\figwidth]{{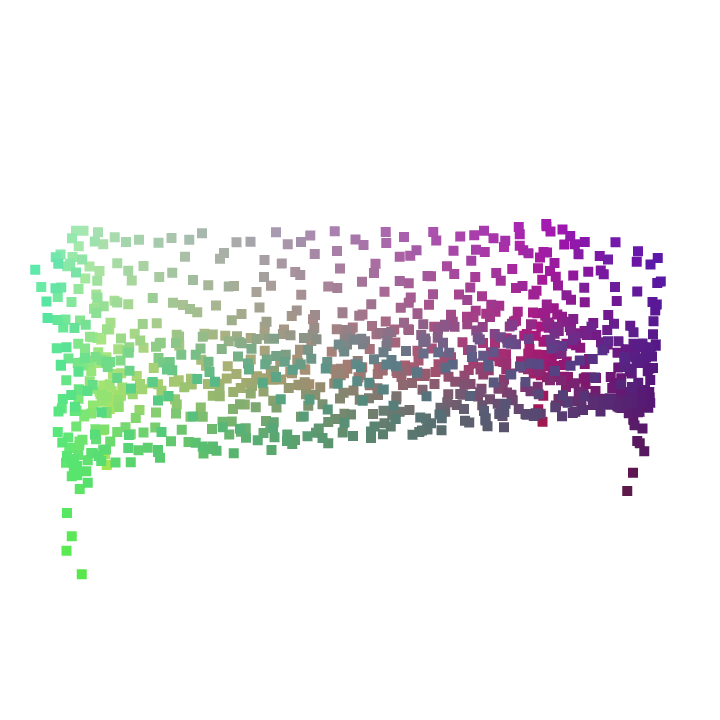}} &
\includegraphics[width=\figwidth]{{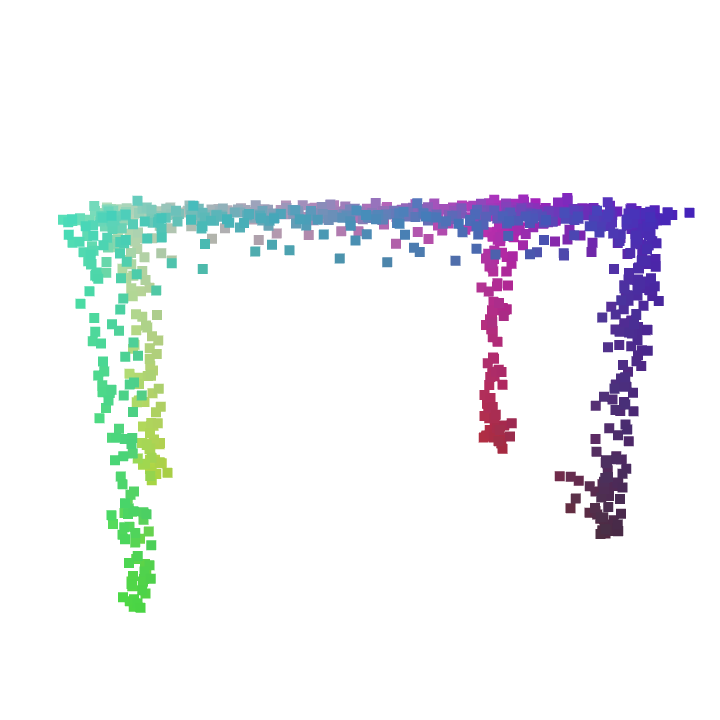}} &
\includegraphics[width=\figwidth]{{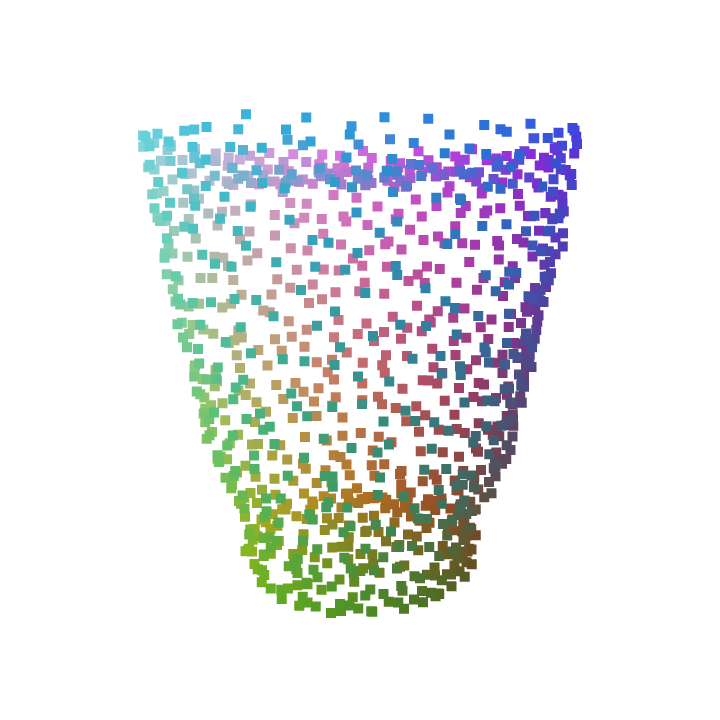}} \cr
PointAE-AtlasNet &
\includegraphics[width=\figwidth]{{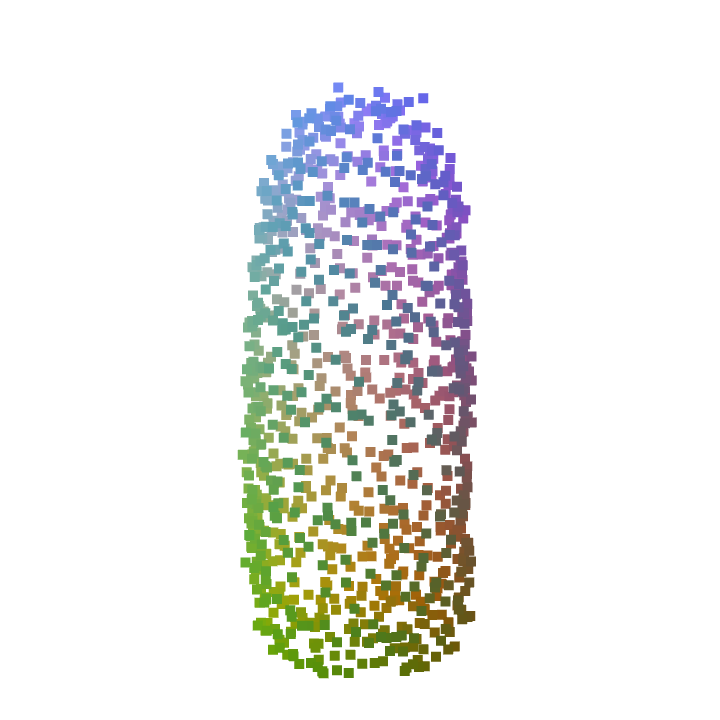}} &
\includegraphics[width=\figwidth]{{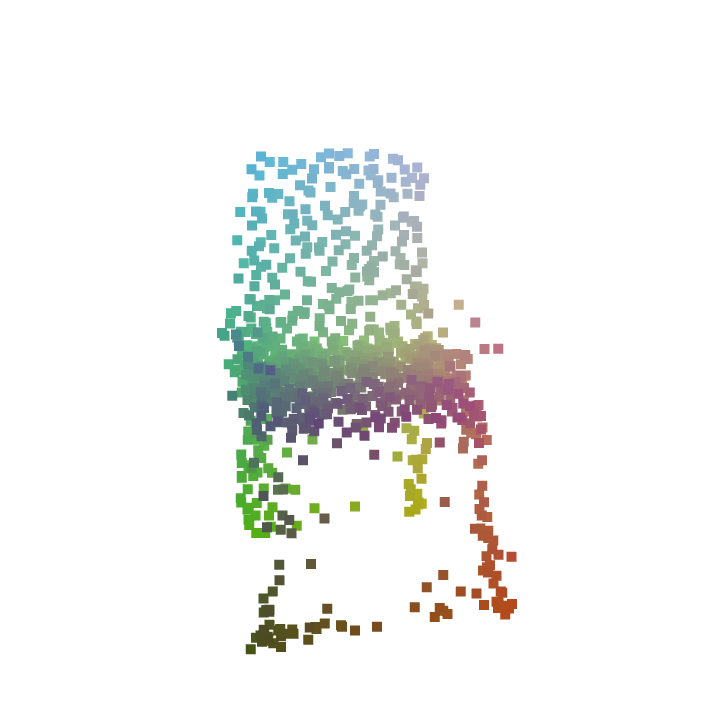}} &
\includegraphics[width=\figwidth]{{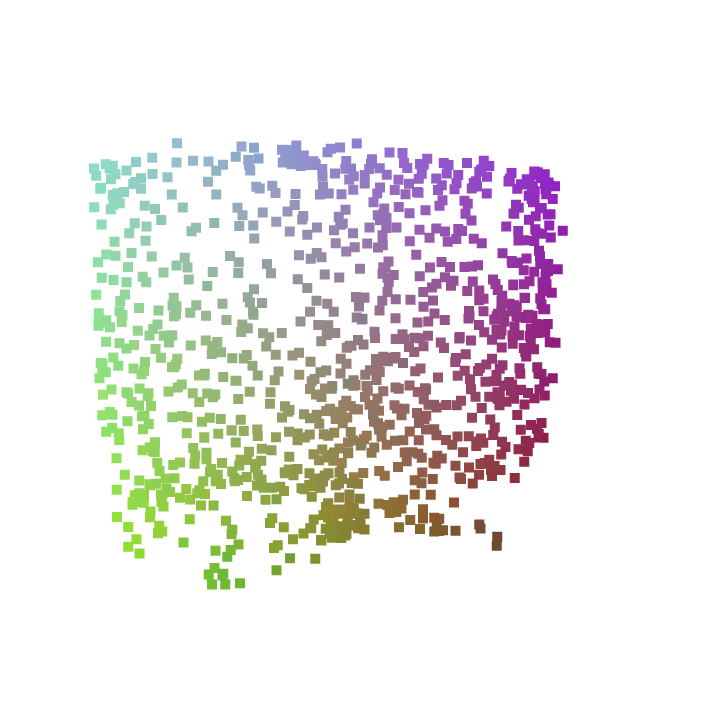}} &
\includegraphics[width=\figwidth]{{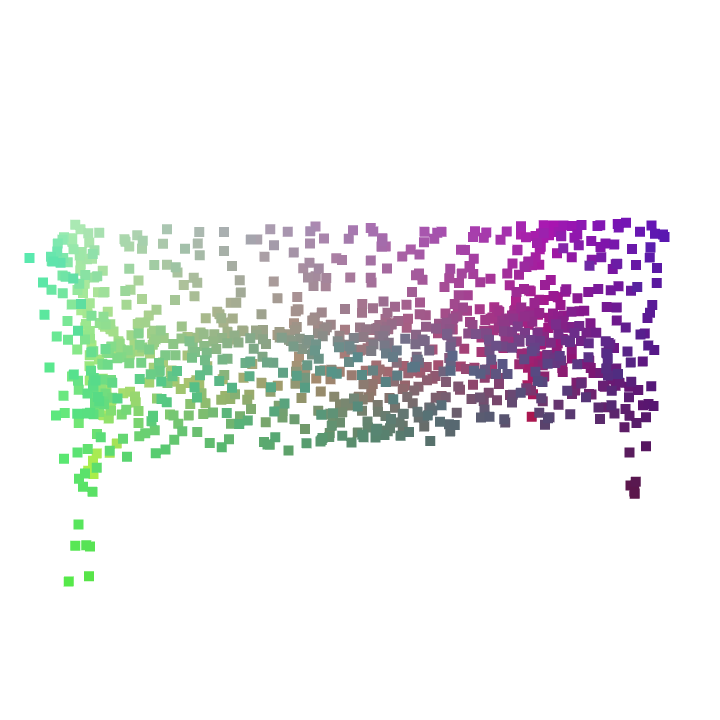}} &
\includegraphics[width=\figwidth]{{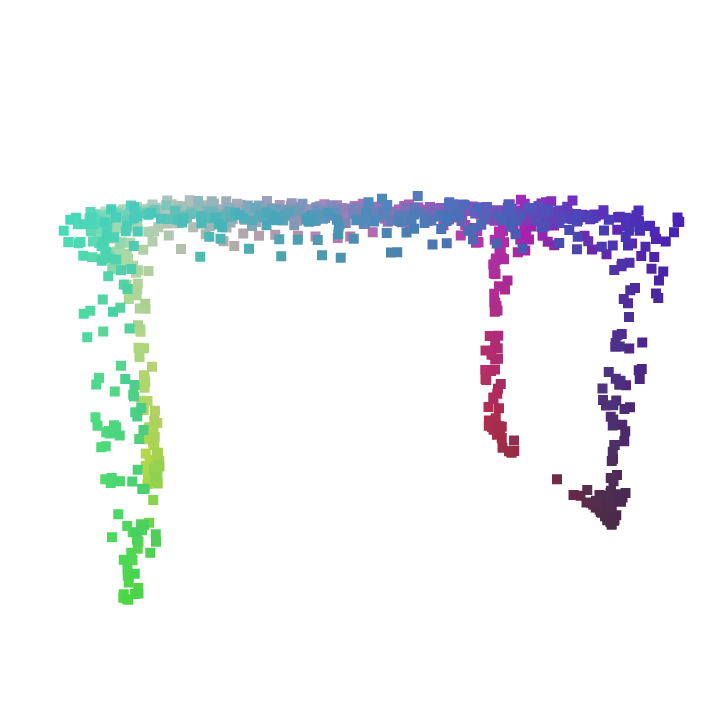}} &
\includegraphics[width=\figwidth]{{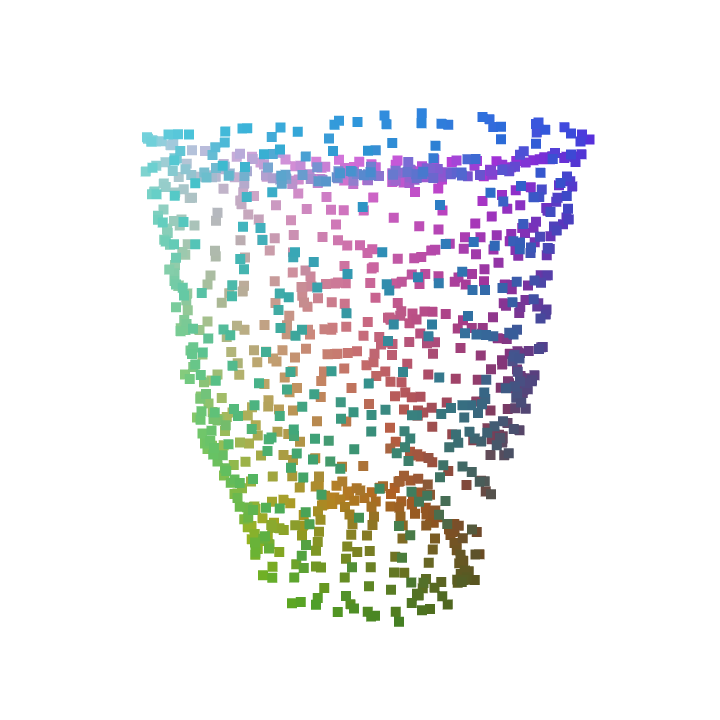}} \cr
\end{tabular}
\cutcaptionup
\caption{Qualitative results of ShapeAdv Auxiliary under \textbf{untargeted} attacks. The adversaries resemble both the input and the auxiliary point clouds.}
\cutcaptiondown
\label{fig:aux}
\end{figure}

\cutsubsectionup
\subsection{Experimental Results}
\cutsubsectiondown

\cutparagraphup
\paragraph{Overall analysis: shape-aware adversarial attacks.}
\cutparagraphdown
First, we measure the attack success rate of our ShapeAdv methods under both targeted and untargeted settings, and found that
the attack success rate is 100\% for all methods.
We report the perturbation magnitude in the Chamfer distance in Table~\ref{tab:exp-shapeadv-all}.
Overall, adversarial point clouds generated by the AtlasNet~\cite{groueix2018atlasnet} have relatively small perturbations compared to the MLP baseline~\cite{achlioptas2018learning}.
Compared to the other variants, our ShapeAdv-Chamfer achieves the lowest distance, as it is directly used in the optimization objective when generating the adversarial attacks.
In particular, our ShapeAdv-Auxiliary generates attacks that deviate further from the original shape as we use one more auxiliary point cloud to guide the shape deformation, such that the distance to the original shape is not always minimized.
As in Table~\ref{tab:exp-shapeadv-all}, the trend is consistent in both targeted and untargeted settings, while untargeted attacks are relatively easier to accomplish than targeted ones.

In Figure~\ref{fig:autoencoder}, we visualize our adversarial examples for targeted attacks.
%
Both methods generate perceptually plausible 3D point clouds similar to the shape of the original one with noticeable shape deformations.
AtlasNet tends to preserve local geometric details of the original point cloud; the MLP baseline tends to generate a coarse shape of the same semantic category while being slightly different from the original point cloud.
Compared to ShapeAdv-Chamfer which uses the Chamfer distance objective, examples generated by ShapeAdv-Latent~$\ell_2$ have relatively larger
deformations, which is consistent with the distance measure in our quantitative evaluation.
Taking account of its better visualization, we use AtlasNet for ShapeAdv below, unless otherwise stated.

\cutparagraphup
\paragraph{Deformation attacks with auxiliary point clouds.}
\cutparagraphdown
As seen in Figure~\ref{fig:aux}, the deformed shapes guided by auxiliary point clouds under adversarial attacks are perceptually similar to both input and auxiliary point cloud in one or several aspects.
Moreover, this attack works in a more controllable fashion in which the deformation can be guided by the geometry and structure of auxiliary point cloud.
For example, in the third column of Figure~\ref{fig:aux}, the method is able to generate an adversarial chair without chair arms from the input chair with two arms.
This can be potentially useful when analyzing the robustness of modern 3D point cloud classifiers given a specific shape deformation.

\begin{figure}[!t]
\centering
\scriptsize\setlength{\tabcolsep}{0mm}
\begin{tabular}{
>{\centering}m{14mm}
>{\centering}m{\figwidth}
>{\centering}m{\figwidth}
>{\centering}m{\figwidth}
>{\centering}m{\figwidth}
>{\centering}m{\figwidth}
>{\centering}m{\figwidth}
}
\multirow{2}{*}{\parbox{14mm}{\centering Attack Method}} & \multicolumn{2}{c}{Airplane $\rightarrow$ Bottle} &
\multicolumn{2}{c}{Bottle $\rightarrow$ Chair} &
\multicolumn{2}{c}{Chair $\rightarrow$ Airplane} \cr
& Front & Side & Front & Top & Front & Side \cr
\hline
No Attack &
\includegraphics[width=\figwidth]{{figs/comp/clean_0_5_13_orig_front.png}} &
\includegraphics[width=\figwidth]{{figs/comp/clean_0_5_13_orig_side.png}} &
\includegraphics[width=\figwidth]{{figs/comp/clean_5_8_14_orig_front.png}} &
\includegraphics[width=\figwidth]{{figs/comp/clean_5_8_14_orig_side.png}} &
\includegraphics[width=\figwidth]{{figs/comp/clean_8_0_12_orig_front.png}} &
\includegraphics[width=\figwidth]{{figs/comp/clean_8_0_12_orig_side.png}} \cr
\hline
Shift-Point~\cite{xiang2019generating} &
\includegraphics[width=\figwidth]{{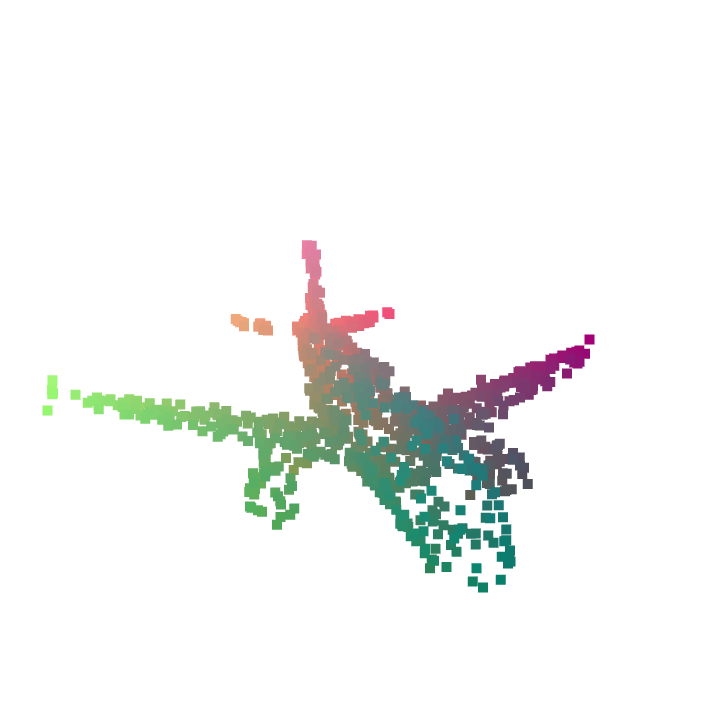}} &
\includegraphics[width=\figwidth]{{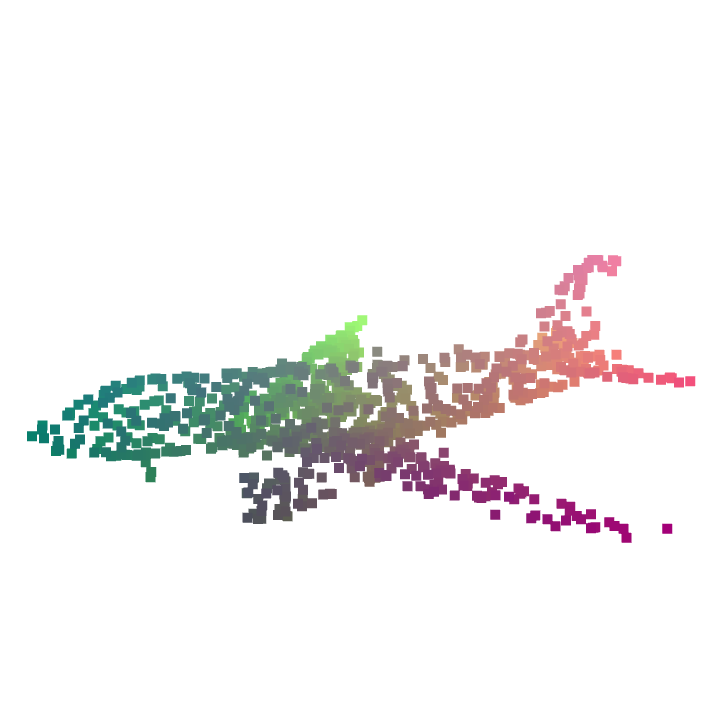}} &
\includegraphics[width=\figwidth]{{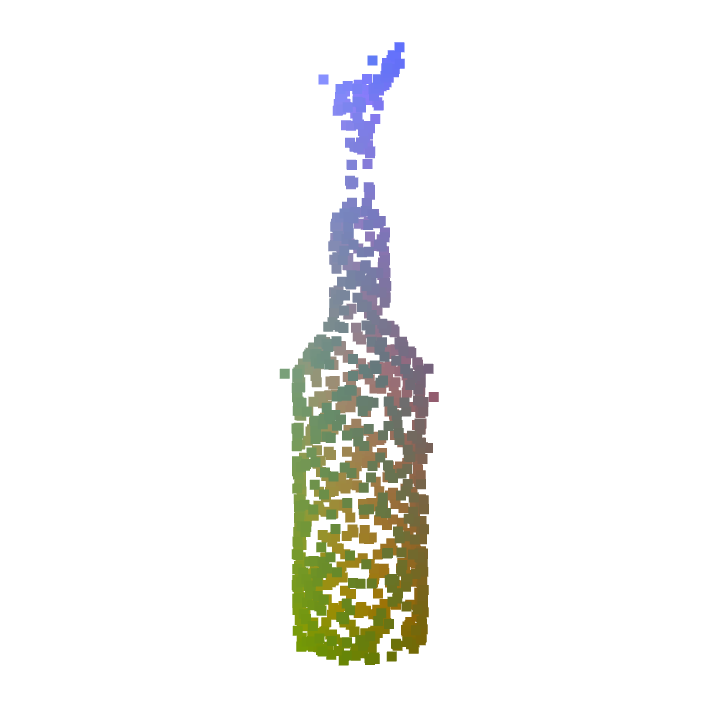}} &
\includegraphics[width=\figwidth]{{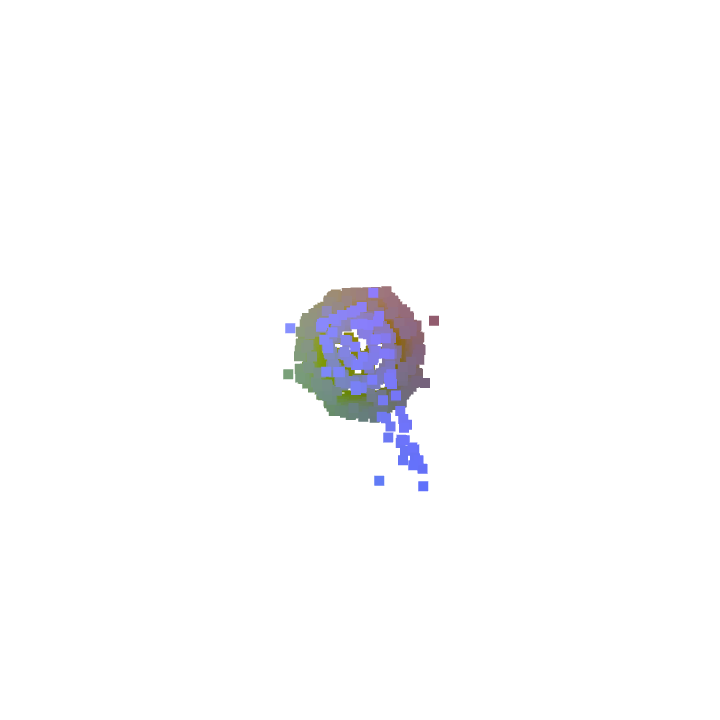}} &
\includegraphics[width=\figwidth]{{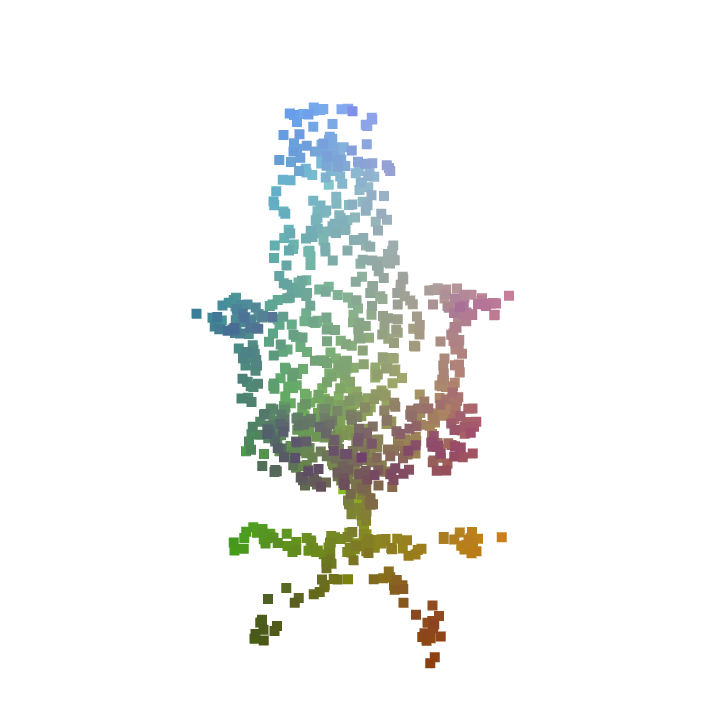}} &
\includegraphics[width=\figwidth]{{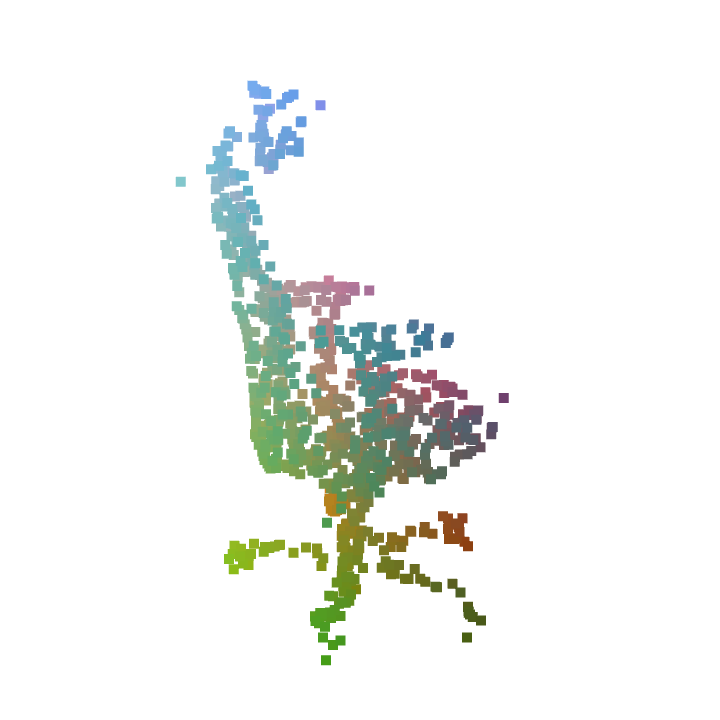}} \cr
Add-Point~\cite{xiang2019generating} &
\includegraphics[width=\figwidth]{{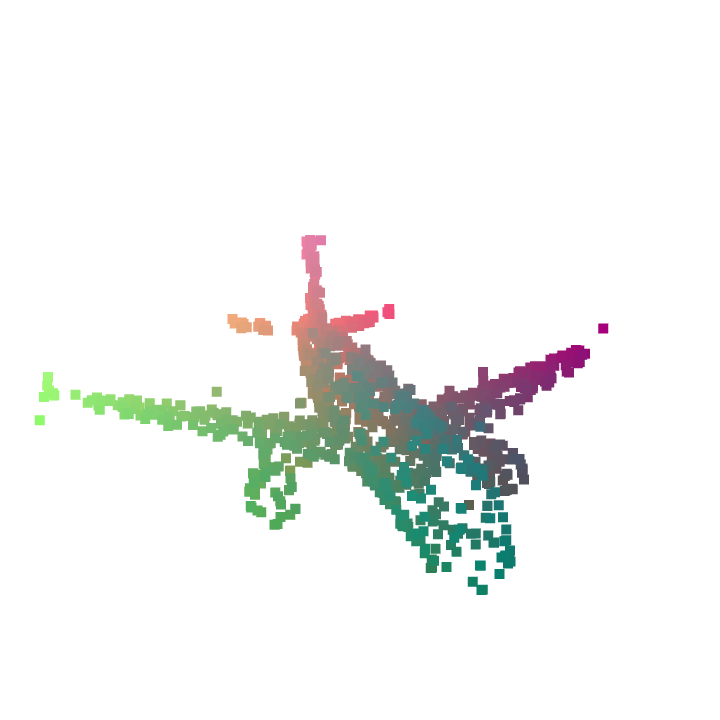}} &
\includegraphics[width=\figwidth]{{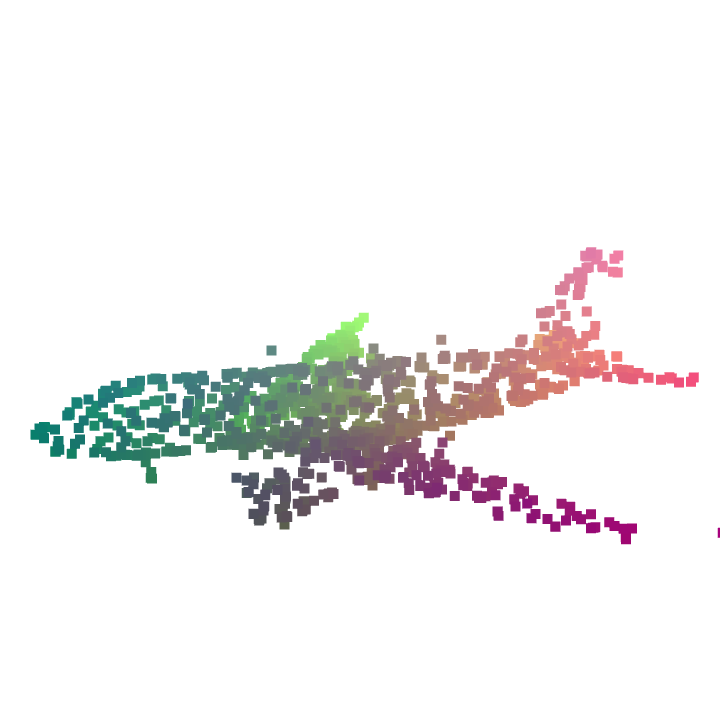}} &
\includegraphics[width=\figwidth]{{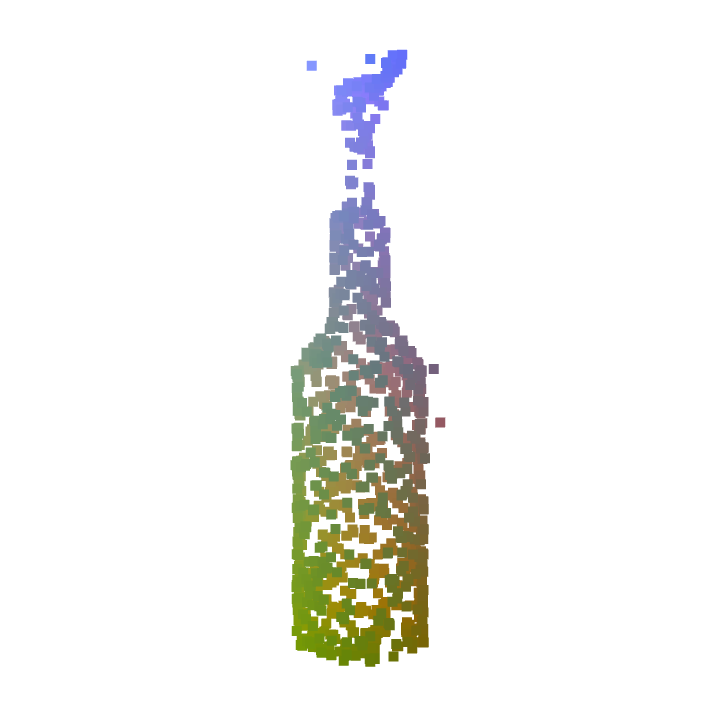}} &
\includegraphics[width=\figwidth]{{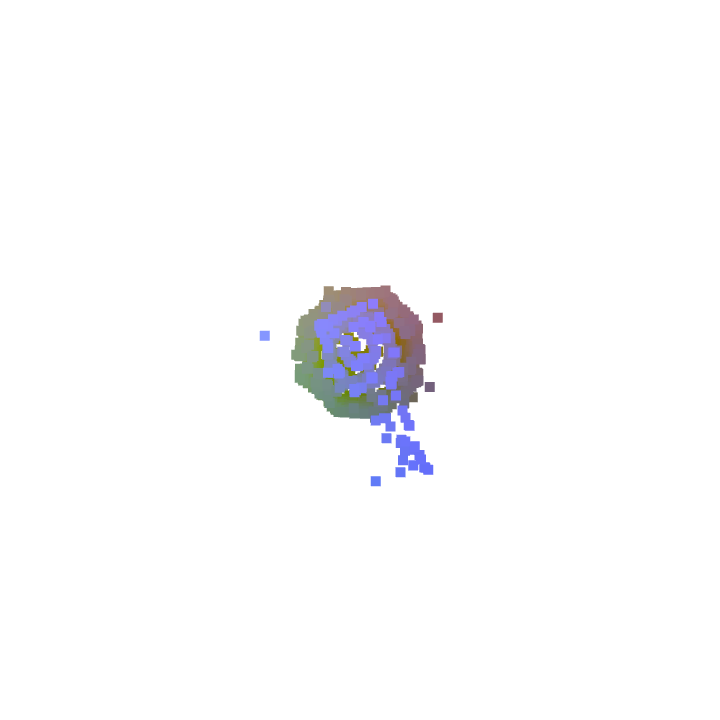}} &
\includegraphics[width=\figwidth]{{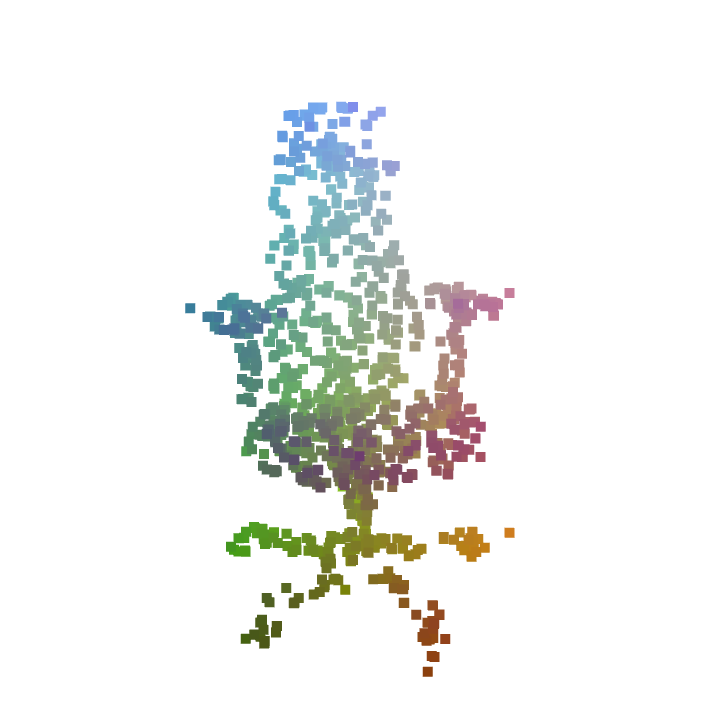}} &
\includegraphics[width=\figwidth]{{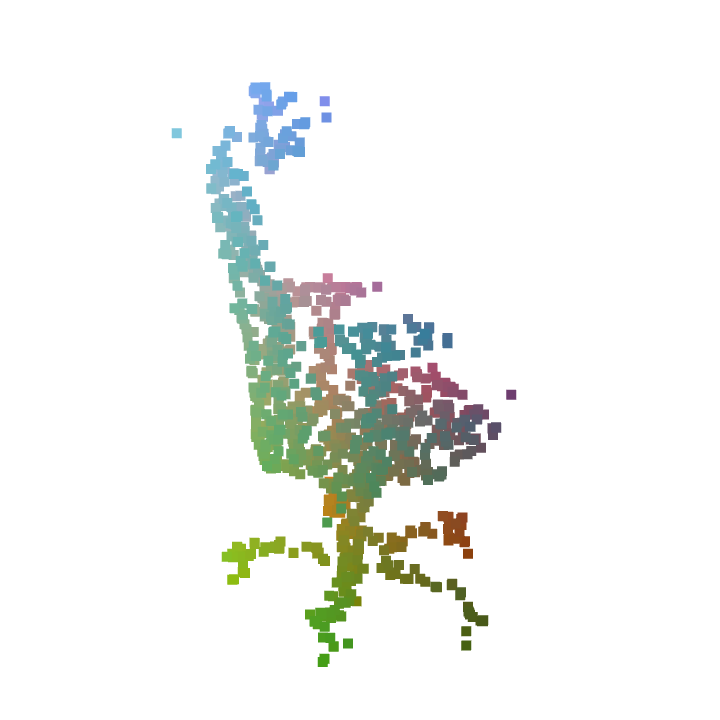}} \cr
\hline
Add-Cluster~\cite{xiang2019generating} &
\includegraphics[width=\figwidth]{{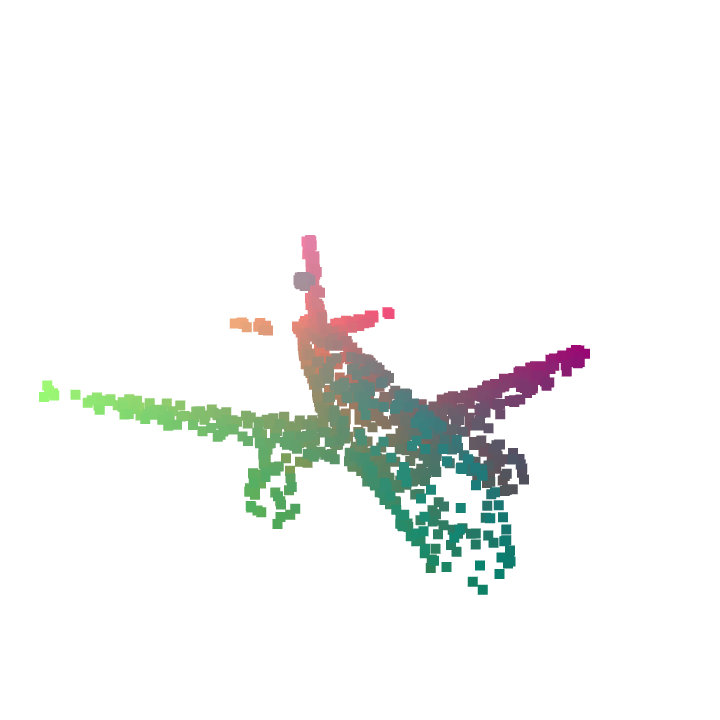}} &
\includegraphics[width=\figwidth]{{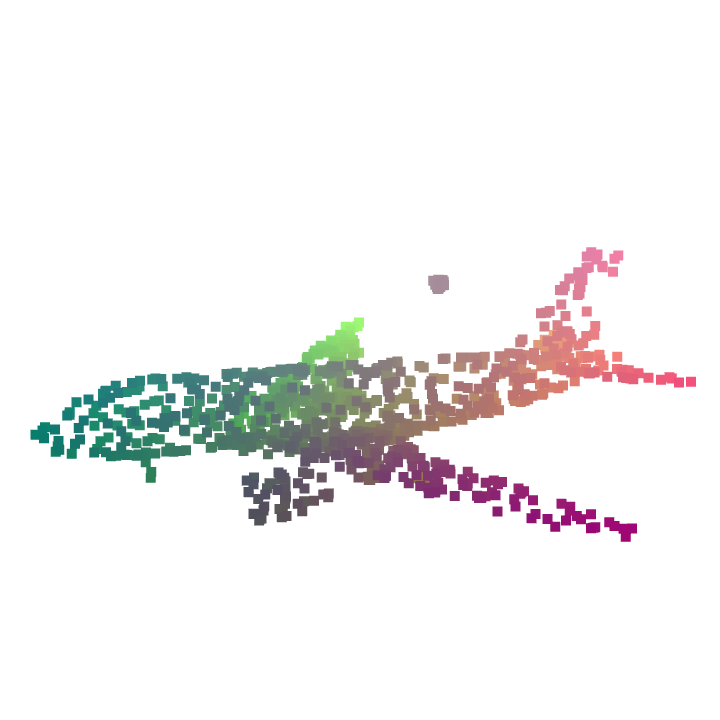}} &
\includegraphics[width=\figwidth]{{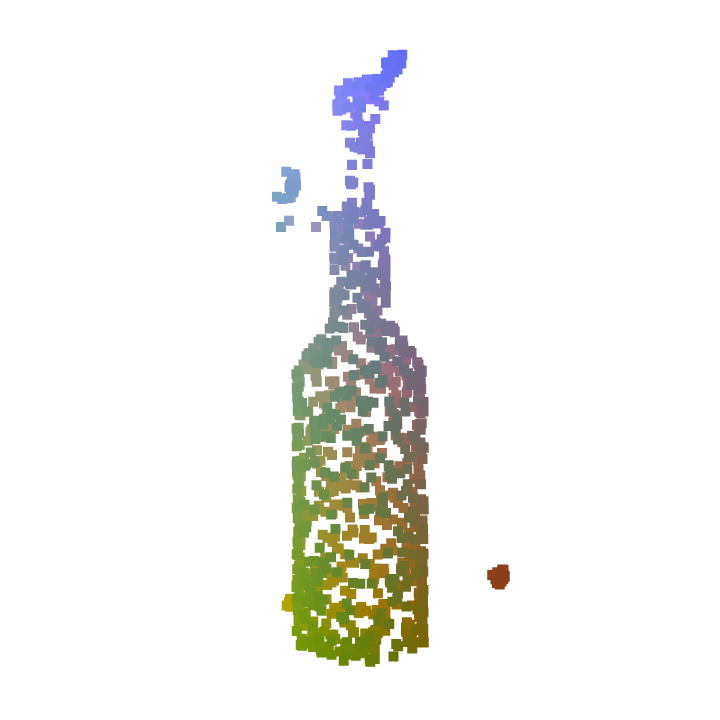}} &
\includegraphics[width=\figwidth]{{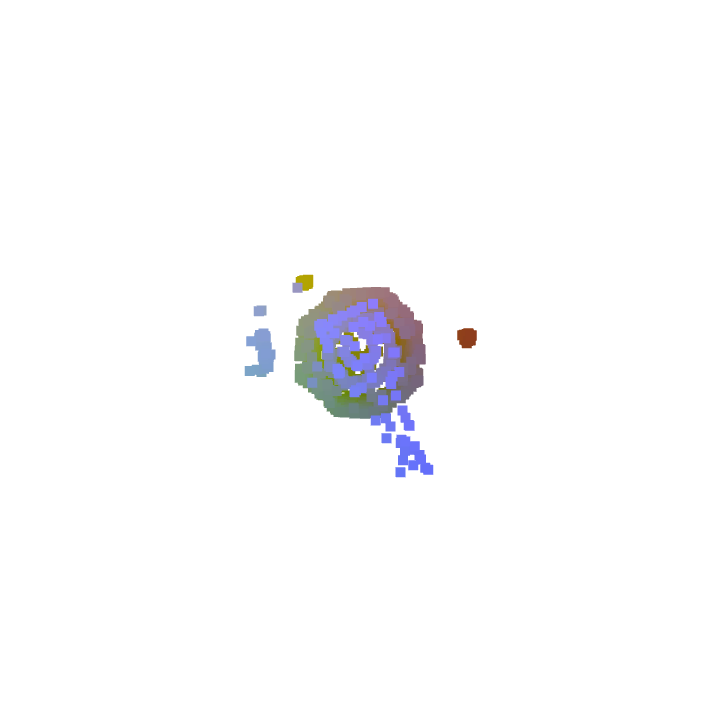}} &
\includegraphics[width=\figwidth]{{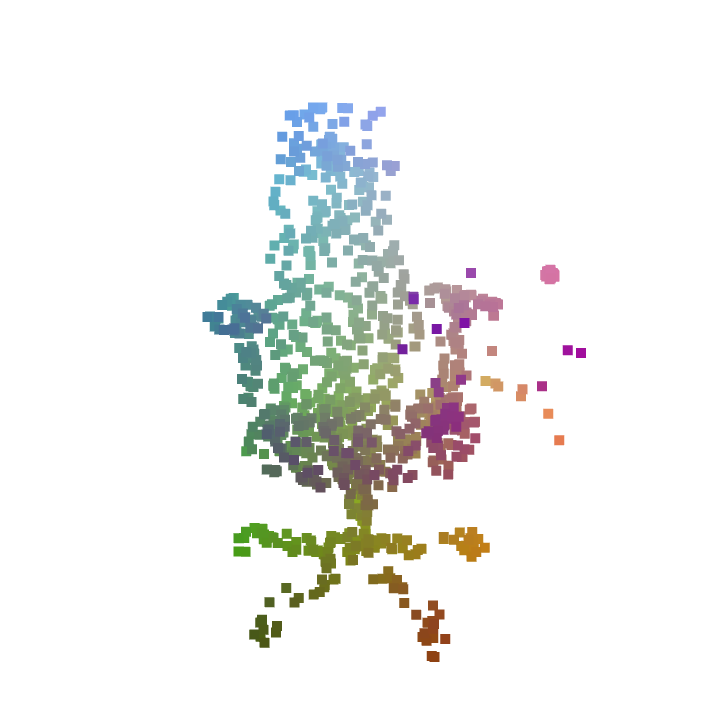}} &
\includegraphics[width=\figwidth]{{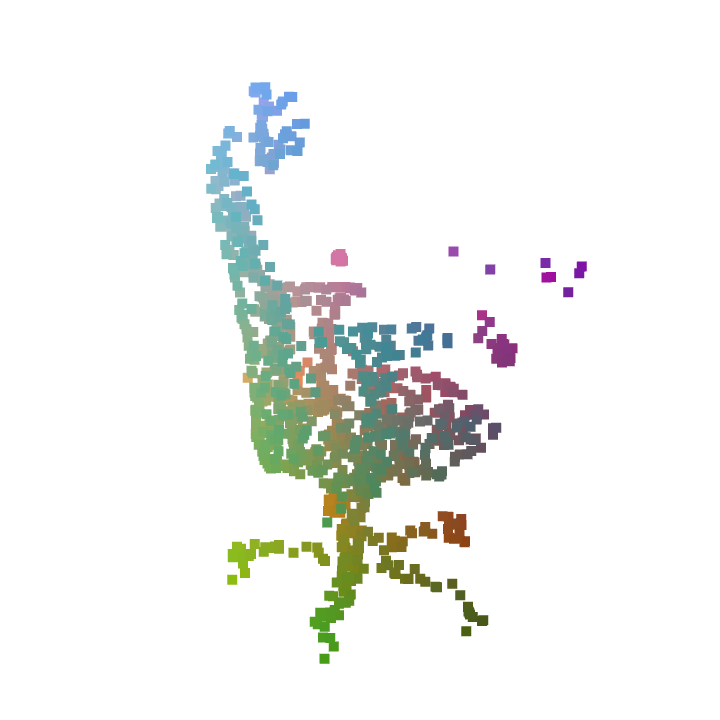}} \cr
Add-Object~\cite{xiang2019generating} &
\includegraphics[width=\figwidth]{{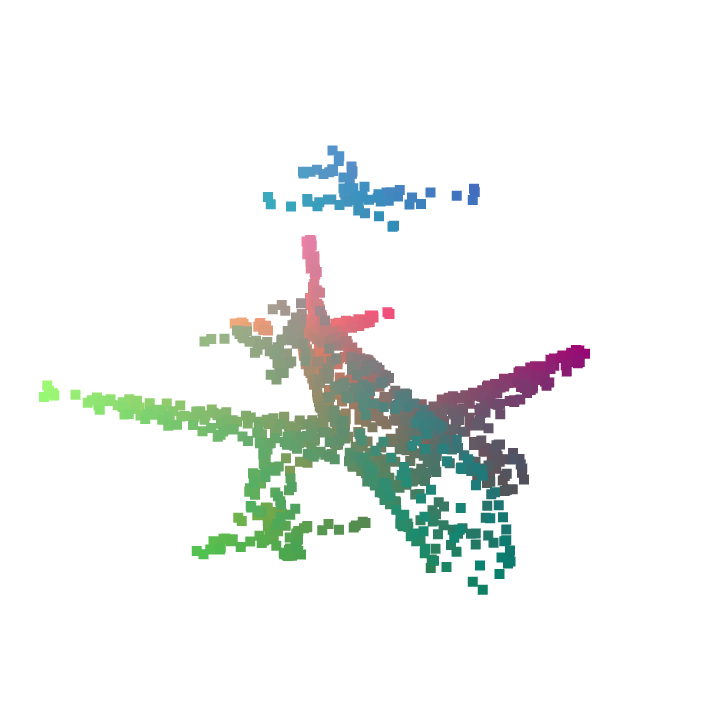}} &
\includegraphics[width=\figwidth]{{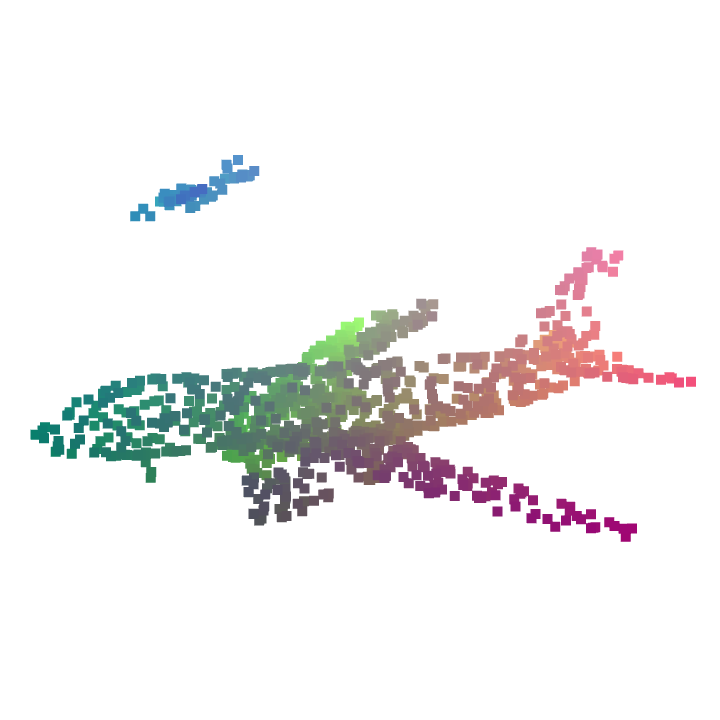}} &
\includegraphics[width=\figwidth]{{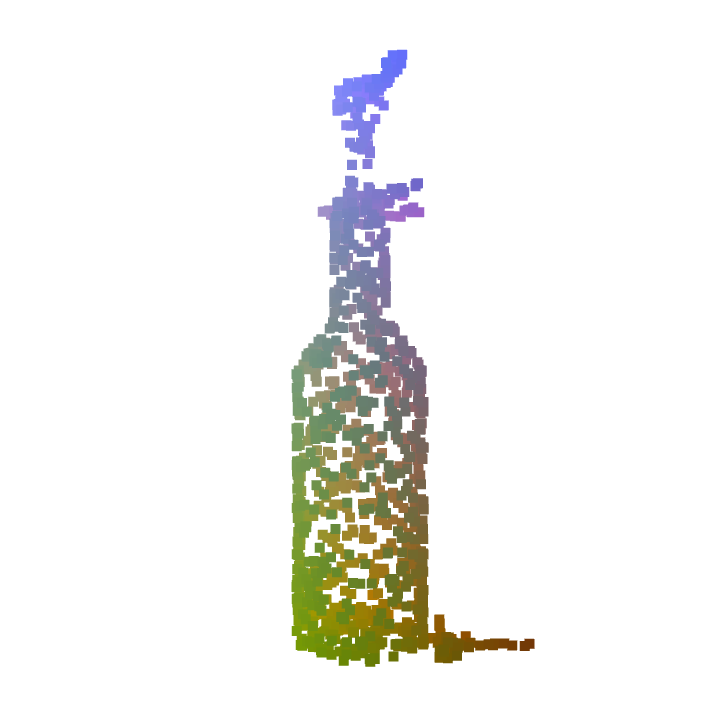}} &
\includegraphics[width=\figwidth]{{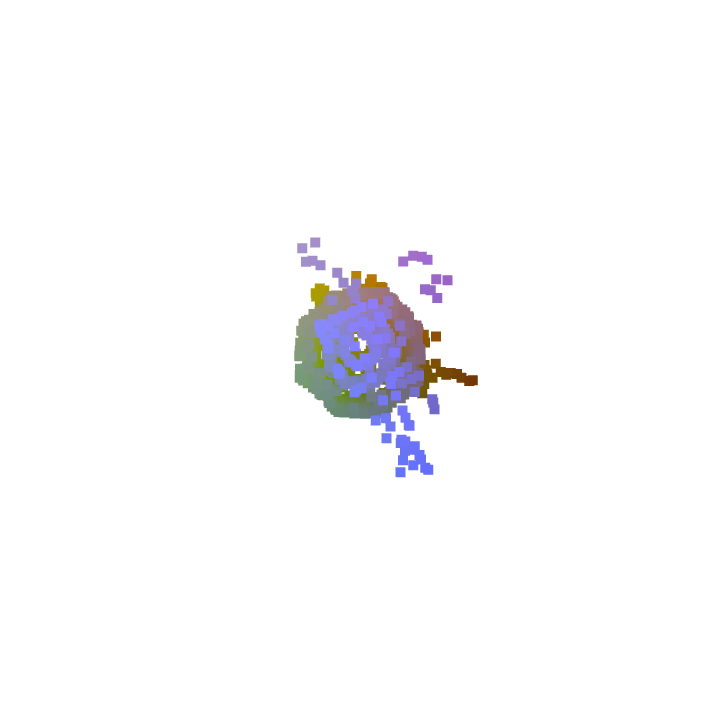}} &
\includegraphics[width=\figwidth]{{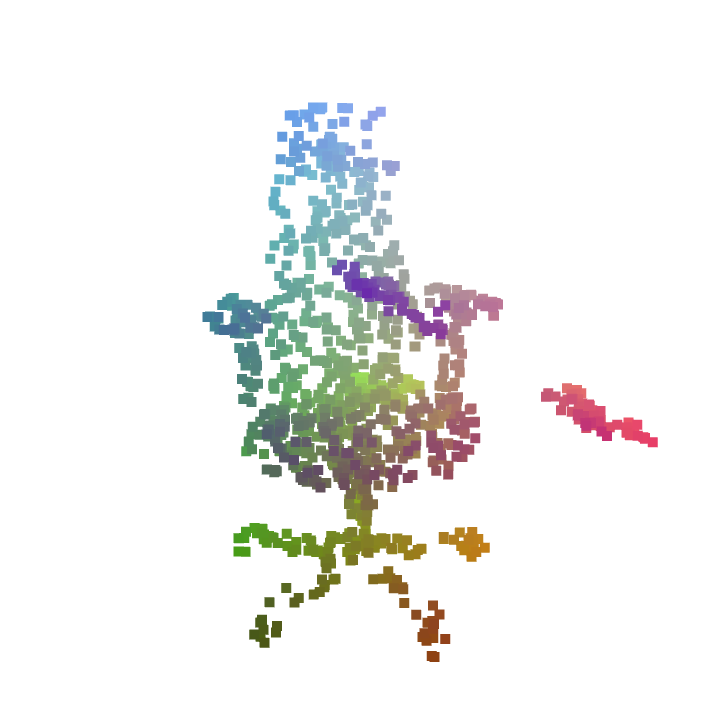}} &
\includegraphics[width=\figwidth]{{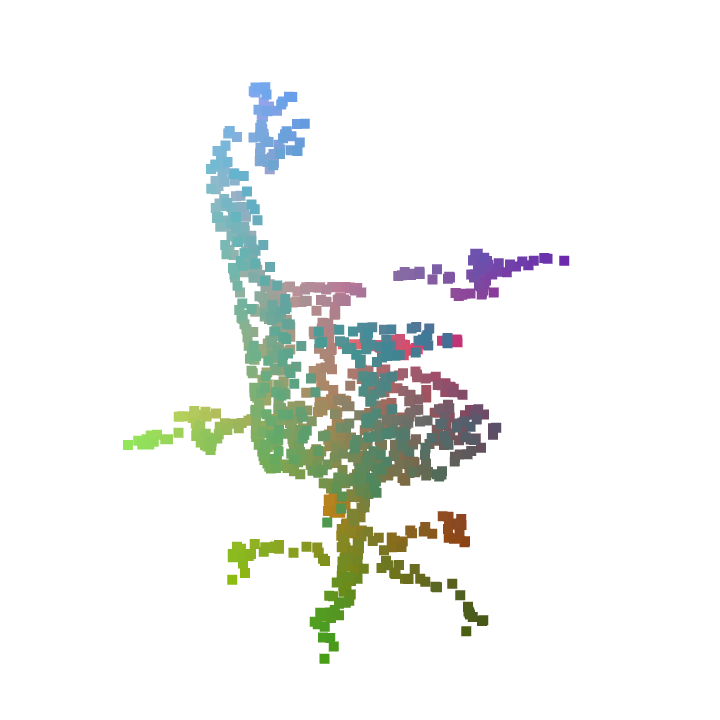}} \cr
\hline
ShapeAdv Latent~$\ell_2$ (Ours) &
\includegraphics[width=\figwidth]{{figs/comp/latentl2_atlas_0_5_13_adv_front.png}} &
\includegraphics[width=\figwidth]{{figs/comp/latentl2_atlas_0_5_13_adv_side.png}} &
\includegraphics[width=\figwidth]{{figs/comp/latentl2_atlas_5_8_14_adv_front.png}} &
\includegraphics[width=\figwidth]{{figs/comp/latentl2_atlas_5_8_14_adv_side.png}} &
\includegraphics[width=\figwidth]{{figs/comp/latentl2_atlas_8_0_12_adv_front.png}} &
\includegraphics[width=\figwidth]{{figs/comp/latentl2_atlas_8_0_12_adv_side.png}} \cr
ShapeAdv Chamfer (Ours) &
\includegraphics[width=\figwidth]{{figs/comp/ch_atlas_0_5_13_adv_front.png}} &
\includegraphics[width=\figwidth]{{figs/comp/ch_atlas_0_5_13_adv_side.png}} &
\includegraphics[width=\figwidth]{{figs/comp/ch_atlas_5_8_14_adv_front.png}} &
\includegraphics[width=\figwidth]{{figs/comp/ch_atlas_5_8_14_adv_side.png}} &
\includegraphics[width=\figwidth]{{figs/comp/ch_atlas_8_0_12_adv_front.png}} &
\includegraphics[width=\figwidth]{{figs/comp/ch_atlas_8_0_12_adv_side.png}} \cr
\end{tabular}
\cutcaptionup
\caption{Qualitative comparison of ShapeAdv and prior works under \textbf{targeted} attacks. ShapeAdv methods generate perceptually plausible point clouds without introducing significant outliers.
%
%
%
%
}
\cutcaptiondown
\label{fig:comparison}
\end{figure}

\vspace*{-10pt}

\begin{table}[t]
	\cuttableup
	\centering
	\scalebox{0.8}{
	\begin{tabular}{l||c|c|c||c|c|c|c}
		\toprule
		Evaluation Metric & \multicolumn{3}{c||}{Attack Success Rate (\%) ($\uparrow$)} & \multicolumn{4}{c}{Chamfer distance ($\times 10^{-3}$) ($\downarrow$)} \cr
		\midrule
		Attack / Defense &
		SOR~\cite{rusu2008towards} & AE-MLP & AE-AtlasNet &
		No Def. & SOR~\cite{rusu2008towards} & AE-MLP & AE-AtlasNet \cr
		\midrule
		Shift-Point~\cite{xiang2019generating} &
		 \textbf{10.0} & 9.6 & 9.5 &
		 0.15 & 0.78 & \textbf{4.51} & \textbf{3.91} \cr
		Add-Point~\cite{xiang2019generating} &
		 7.8 & \textbf{10.3} & \textbf{9.8} &
		 \textbf{0.09} & \textbf{0.62} & 5.13 & 4.20 \cr
		\midrule
		Add-Cluster~\cite{xiang2019generating} &
		 63.5 & \textbf{53.2} & \textbf{34.0} &
		 17.60 & 19.21 & 22.23 & 12.42 \cr
		Add-Object~\cite{xiang2019generating} &
		 \textbf{68.3} & 37.1 & 31.0 &
		 \textbf{12.16} & \textbf{13.86} & \textbf{15.36} & \textbf{10.81} \cr
		\midrule
		MLP-Latent~$\ell_2$ (Eq.~\eqref{eqn:shapeadv_l2}) &
		\textbf{34.4} & \textbf{24.7} & \textbf{23.4} &
		8.56 & 9.60 & 9.84 & 9.37 \cr
		MLP-Chamfer  (Eq.~\eqref{eqn:shapeadv_ch}) &
		20.8 & 14.6 & 15.1 &
		4.55 & 5.63 & 5.74 & 5.59 \cr
	    Atlas-Latent~$\ell_2$ (Eq.~\eqref{eqn:shapeadv_l2}) &
		 15.8 & 11.2 & 12.7 &
		 4.29 & 4.82 & 5.59 & 5.02 \cr
		Atlas-Chamfer (Eq.~\eqref{eqn:shapeadv_ch}) &
		 15.4 & 10.8 & 10.4 &
		 \textbf{3.15} & \textbf{3.90} & \textbf{5.03} & \textbf{4.38} \cr
		\bottomrule
	\end{tabular}
	}
	\cuttablecaptionup
	\caption{
		Quantitative evaluation of \textbf{targeted} attacks in terms of attack success rate (\%) and the Chamfer distance ($\times 10^{-3}$).
		$\uparrow$ ($\downarrow$) implies that the higher (lower) number is the better.
		We compare our shape-aware attacks with existing attacks against SOR~\cite{rusu2008towards,zhou2019dup} and our AE defenses extended from Defense-GAN~\cite{samangouei2018defense}.
		%
	}
    \cuttabledown
    \cuttabledown
	\label{tab:target-att-def-table}
\end{table}

\cutparagraphup
\paragraph{Shape-aware adversarial attacks against defense methods.}
\cutparagraphdown
We compare our shape-aware attacks with existing point cloud attack methods proposed by \cite{xiang2019generating} and visualize the results in Figure~\ref{fig:comparison}.
We evaluate the proposed shape-aware attacks against two defense methods including sparse outlier removal (SOR)~\cite{zhou2019dup} and our auto-encoder based defense method, which can also be interpreted as Defense-GAN~\cite{samangouei2018defense} on 3D point clouds.
SOR is designed to remove point outliers from the raw point cloud while our AE-based method performs the defense through the auto-encoder.
As in Table~\ref{tab:target-att-def-table}, our shape-aware adversarial attacks are generally more difficult to defend than the existing point perturbation attacks and attacks based on adding individual points.
Compared to them, adversarial point clouds generated by our shape-aware attacks have noticeable geometric deformations as we do not use point-to-point correspondences to constrain the perturbation.
Also, our methods exhibit different statistics compared to attacks by adding points from point clusters or objects, as the resulting shapes from those methods are severely deformed from the origin ones (the Chamfer distance is higher).
In summary, we believe that our shape-aware adversarial attacks are orthogonal to existing adversarial attacks on point clouds.

Comparisons between the two architectures (MLP and AtlasNet) demonstrate that ShapeAdv generated by the MLP baseline is relatively harder to defend against compared to AtlasNet.
First, this implies that AtlasNet tends to inject adversarial perturbations on the local geometry surfaces which can be partially removed by the existing defense methods, while the MLP baseline tends to inject global deformations as adversarial perturbations which makes the defense challenging.
%
%
When it comes to 3D point cloud reconstruction, AtlasNet is more expressive and detail-preserving with a patch-based decoder that constructs the piecewise planar surfaces, where each planar surface is independent from each other (each patch is generated by a separate deformation-based decoder with different model weights).
However, the detail-preserving adversaries can be possibly considered as outliers by the point defense methods.
Second, existing defense methods consistently fail at defending against adversaries with global geometric deformation or structure difference while being effective against adversary with only local geometric deformation.

\vspace*{-3pt}
\cutparagraphup
\paragraph{Shape-aware attack transferability.}
\cutparagraphdown
Regarding the safety-critical concerns related to the PointNet model~\cite{qi2017pointnet}, we further extend our study on black-box attack transferability.
Basically, we study how well the attacks generated by our shape-aware attack methods can be transferred to other classifiers such as PointNet++~\cite{qi2017pointnet++} and DGCNN~\cite{wang2018dynamic}.
As in Table~\ref{tab:untarget-transfer}, our latent-space $\ell_2$ attacks exhibit stronger transferability compared to all existing methods.
Here, we exclude the comparisons with methods that generate additional clusters or objects as they lead to severe shape deformation to the original object shapes.

\begin{table}[t]
	\centering
	\setlength{\tabcolsep}{4pt} 
	\cuttableup
	\scalebox{0.9}{
	\begin{tabular}{l|c|c|c}
		\toprule
		& PointNet++ & DGCNN & PointNet$^{*}$ \cr
		\midrule
		Shift-Point~\cite{xiang2019generating} &  3.9\% & 1.9\% & 5.5\%\cr
		Add-Point~\cite{xiang2019generating} & 3.6\% & 7.4\% & 7.8\%\cr
		\midrule
		MLP-Latent~$\ell_2$ (Eq.~\eqref{eqn:shapeadv_l2}) & \textbf{24.7\%} & \textbf{23.5\%} & \textbf{13.9\%}\cr
		MLP-Chamfer (Eq.~\eqref{eqn:shapeadv_ch}) & \textbf{16.6\%} & \textbf{17.4\%} & 10.8\% \cr
		MLP-Auxiliary (Eq.~\eqref{eqn:shapeadv_aux})& 13.5\% &14.8\% &6.5\%\cr
		\midrule
		Atlas-Latent~$\ell_2$ (Eq.~\eqref{eqn:shapeadv_l2}) &  13.6\% & \textbf{14.2\%} & \textbf{11.6\%}\cr
		Atlas-Chamfer (Eq.~\eqref{eqn:shapeadv_ch}) & \textbf{13.9\%} &13.0\% &11.0\% \cr
		Atlas-Auxiliary (Eq.~\eqref{eqn:shapeadv_aux})& 10.8\% &11.8\% &7.7\%\cr
		\bottomrule
	\end{tabular}
	}
	\cuttablecaptionup
	\caption{Attack success rates of \textbf{untargeted} transfer attacks against PointNet++~\cite{qi2017pointnet++}, DGCNN~\cite{wang2018dynamic}, augmented PointNet, and PointNet with a different weight initialization (PointNet$^{*}$).}
	\cuttabledown
	\cuttabledown
	\label{tab:untarget-transfer}
\end{table}


\vspace*{-2pt}
\cutsectionup
\section{Conclusion}
\cutsectiondown
\vspace*{-2pt}
In this paper, we have studied the robustness of 3D point cloud classifiers 
by exploiting the shape-aware adversarial attacks.
In particular, we proposed to inject adversarial perturbations to the learned latent space of a point auto-encoder 
as an approximation to the shape manifold.
By introducing shape deformations in the latent space, we are able to explore shape variations of a certain class 
in the adversarial point cloud generation.
%
We then extended our shape-aware attacks by guiding the shape deformation with auxiliary point clouds, 
which reflects the process of shape morphing in the latent space.
Moreoever, we have shown that the learned latent representation is directly applicable to a defense method 
against perturbations away from it.
%
We believe our shape-aware adversarial attacks are orthogonal to existing attacks generated directly on the point cloud, 
which can broaden the landscape of adversarial robustness on 3D point cloud.
Besides 3D point cloud representation, the idea of generating adversary on the latent space can potentially be applicable 
to other 3D data representations, such as voxel grid and surface mesh.

	\clearpage
	
	\bibliographystyle{splncs04}
	\bibliography{reference}

\begin{thebibliography}{10}
\providecommand{\url}[1]{\texttt{#1}}
\providecommand{\urlprefix}{URL }
\providecommand{\doi}[1]{https://doi.org/#1}

\bibitem{achlioptas2018learning}
Achlioptas, P., Diamanti, O., Mitliagkas, I., Guibas, L.: Learning
  representations and generative models for 3d point clouds. In: ICML (2018)

\bibitem{alzantot2018generating}
Alzantot, M., Sharma, Y., Elgohary, A., Ho, B.J., Srivastava, M., Chang, K.W.:
  Generating natural language adversarial examples. arXiv preprint
  arXiv:1804.07998  (2018)

\bibitem{bhattad2019big}
Bhattad, A., Chong, M.J., Liang, K., Li, B., Forsyth, D.A.: Big but
  imperceptible adversarial perturbations via semantic manipulation. ICLR
  (2020)

\bibitem{blanz1999morphable}
Blanz, V., Vetter, T.: A morphable model for the synthesis of 3d faces. In: ACM
  SIGGRAPH. vol.~99, pp. 187--194 (1999)

\bibitem{bogo2016keep}
Bogo, F., Kanazawa, A., Lassner, C., Gehler, P., Romero, J., Black, M.J.: Keep
  it smpl: Automatic estimation of 3d human pose and shape from a single image.
  In: ECCV (2016)

\bibitem{brock2016generative}
Brock, A., Lim, T., Ritchie, J.M., Weston, N.: Generative and discriminative
  voxel modeling with convolutional neural networks. arXiv preprint
  arXiv:1608.04236  (2016)

\bibitem{cao2019adversarial}
Cao, Y., Xiao, C., Cyr, B., Zhou, Y., Park, W., Rampazzi, S., Chen, Q.A., Fu,
  K., Mao, Z.M.: Adversarial sensor attack on lidar-based perception in
  autonomous driving. arXiv preprint arXiv:1907.06826  (2019)

\bibitem{carlini2016defensive}
Carlini, N., Wagner, D.: Defensive distillation is not robust to adversarial
  examples. arXiv preprint arXiv:1607.04311  (2016)

\bibitem{carlini2017towards}
Carlini, N., Wagner, D.: Towards evaluating the robustness of neural networks.
  In: 2017 IEEE Symposium on Security and Privacy (S\&P). IEEE (2017)

\bibitem{chang2015shapenet}
Chang, A.X., Funkhouser, T., Guibas, L., Hanrahan, P., Huang, Q., Li, Z.,
  Savarese, S., Savva, M., Song, S., Su, H., et~al.: Shapenet: An
  information-rich 3d model repository. arXiv preprint arXiv:1512.03012  (2015)

\bibitem{choy20163d}
Choy, C.B., Xu, D., Gwak, J., Chen, K., Savarese, S.: 3d-r2n2: A unified
  approach for single and multi-view 3d object reconstruction. In: ECCV (2016)

\bibitem{fan2017point}
Fan, H., Su, H., Guibas, L.J.: A point set generation network for 3d object
  reconstruction from a single image. In: CVPR (2017)

\bibitem{frosst2018darccc}
Frosst, N., Sabour, S., Hinton, G.: Darccc: Detecting adversaries by
  reconstruction from class conditional capsules. arXiv preprint
  arXiv:1811.06969  (2018)

\bibitem{gadelha2018multiresolution}
Gadelha, M., Wang, R., Maji, S.: Multiresolution tree networks for 3d point
  cloud processing. In: ECCV (2018)

\bibitem{girdhar2016learning}
Girdhar, R., Fouhey, D.F., Rodriguez, M., Gupta, A.: Learning a predictable and
  generative vector representation for objects. In: ECCV (2016)

\bibitem{gkioxari2019mesh}
Gkioxari, G., Malik, J., Johnson, J.: Mesh r-cnn. In: ICCV (2019)

\bibitem{goodfellow2014explaining}
Goodfellow, I.J., Shlens, J., Szegedy, C.: Explaining and harnessing
  adversarial examples. In: ICLR (2015)

\bibitem{groueix2018atlasnet}
Groueix, T., Fisher, M., Kim, V.G., Russell, B.C., Aubry, M.: Atlasnet: A
  papier-m\^ach\'e approach to learning 3d surface generation. In: CVPR (2018)

\bibitem{jalal2017robust}
Jalal, A., Ilyas, A., Daskalakis, C., Dimakis, A.G.: The robust manifold
  defense: Adversarial training using generative models. arXiv preprint
  arXiv:1712.09196  (2017)

\bibitem{jiang2018gal}
Jiang, L., Shi, S., Qi, X., Jia, J.: Gal: Geometric adversarial loss for
  single-view 3d-object reconstruction. In: ECCV (2018)

\bibitem{jin2019ape}
Jin, G., Shen, S., Zhang, D., Dai, F., Zhang, Y.: Ape-gan: Adversarial
  perturbation elimination with gan. In: ICASSP (2019)

\bibitem{joshi2019semantic}
Joshi, A., Mukherjee, A., Sarkar, S., Hegde, C.: Semantic adversarial attacks:
  Parametric transformations that fool deep classifiers. In: ICCV (2019)

\bibitem{kanazawa2018end}
Kanazawa, A., Black, M.J., Jacobs, D.W., Malik, J.: End-to-end recovery of
  human shape and pose. In: CVPR (2018)

\bibitem{landrieu2018large}
Landrieu, L., Simonovsky, M.: Large-scale point cloud semantic segmentation
  with superpoint graphs. In: CVPR (2018)

\bibitem{li2017grass}
Li, J., Xu, K., Chaudhuri, S., Yumer, E., Zhang, H., Guibas, L.: Grass:
  Generative recursive autoencoders for shape structures. ACM Transactions on
  Graphics (TOG)  \textbf{36}(4), ~52 (2017)

\bibitem{liu2019extending}
Liu, D., Yu, R., Su, H.: Extending adversarial attacks and defenses to deep 3d
  point cloud classifiers. In: ICIP (2019)

\bibitem{liu2019soft}
Liu, S., Li, T., Chen, W., Li, H.: Soft rasterizer: A differentiable renderer
  for image-based 3d reasoning. arXiv preprint arXiv:1904.01786  (2019)

\bibitem{mahler2017dex}
Mahler, J., Liang, J., Niyaz, S., Laskey, M., Doan, R., Liu, X., Ojea, J.A.,
  Goldberg, K.: Dex-net 2.0: Deep learning to plan robust grasps with synthetic
  point clouds and analytic grasp metrics. arXiv preprint arXiv:1703.09312
  (2017)

\bibitem{mandikal20183d}
Mandikal, P., Navaneet, K., Agarwal, M., Babu, R.V.: 3d-lmnet: Latent embedding
  matching for accurate and diverse 3d point cloud reconstruction from a single
  image. In: BMVC (2018)

\bibitem{meng2017magnet}
Meng, D., Chen, H.: Magnet: a two-pronged defense against adversarial examples.
  In: Proceedings of the 2017 ACM SIGSAC Conference on Computer and
  Communications Security. pp. 135--147. ACM (2017)

\bibitem{moosavi2016deepfool}
Moosavi-Dezfooli, S.M., Fawzi, A., Frossard, P.: Deepfool: a simple and
  accurate method to fool deep neural networks. In: CVPR (2016)

\bibitem{papernot2017practical}
Papernot, N., McDaniel, P., Goodfellow, I., Jha, S., Celik, Z.B., Swami, A.:
  Practical black-box attacks against machine learning. In: Proceedings of the
  2017 ACM on Asia conference on computer and communications security. pp.
  506--519. ACM (2017)

\bibitem{papernot2016limitations}
Papernot, N., McDaniel, P., Jha, S., Fredrikson, M., Celik, Z.B., Swami, A.:
  The limitations of deep learning in adversarial settings. In: Security and
  Privacy (EuroS\&P), 2016 IEEE European Symposium on (2016)

\bibitem{qi2018frustum}
Qi, C.R., Liu, W., Wu, C., Su, H., Guibas, L.J.: Frustum pointnets for 3d
  object detection from rgb-d data. In: CVPR (2018)

\bibitem{qi2017pointnet}
Qi, C.R., Su, H., Mo, K., Guibas, L.J.: Pointnet: Deep learning on point sets
  for 3d classification and segmentation. In: CVPR (2017)

\bibitem{qi2016volumetric}
Qi, C.R., Su, H., Nie{\ss}ner, M., Dai, A., Yan, M., Guibas, L.J.: Volumetric
  and multi-view cnns for object classification on 3d data. In: CVPR (2016)

\bibitem{qi2017pointnet++}
Qi, C.R., Yi, L., Su, H., Guibas, L.J.: Pointnet++: Deep hierarchical feature
  learning on point sets in a metric space. In: NeurIPS (2017)

\bibitem{qiu2019semanticadv}
Qiu, H., Xiao, C., Yang, L., Yan, X., Lee, H., Li, B.: Semanticadv: Generating
  adversarial examples via attribute-conditional image editing. arXiv preprint
  arXiv:1906.07927  (2019)

\bibitem{ranjan2017improving}
Ranjan, R., Sankaranarayanan, S., Castillo, C.D., Chellappa, R.: Improving
  network robustness against adversarial attacks with compact convolution.
  arXiv preprint arXiv:1712.00699  (2017)

\bibitem{rusu2008towards}
Rusu, R.B., Marton, Z.C., Blodow, N., Dolha, M., Beetz, M.: Towards 3d point
  cloud based object maps for household environments. Robotics and Autonomous
  Systems  \textbf{56}(11),  927--941 (2008)

\bibitem{samangouei2018defense}
Samangouei, P., Kabkab, M., Chellappa, R.: Defense-gan: Protecting classifiers
  against adversarial attacks using generative models. In: ICLR (2018)

\bibitem{schott2019towards}
Schott, L., Rauber, J., Bethge, M., Brendel, W.: Towards the first
  adversarially robust neural network model on mnist. In: ICLR (2019)

\bibitem{sedaghat2017orientation}
Sedaghat, N., Zolfaghari, M., Amiri, E., Brox, T.: Orientation-boosted voxel
  nets for 3d object recognition. In: British Machine Vision Conference (BMVC)
  (2017)

\bibitem{shi2019pointrcnn}
Shi, S., Wang, X., Li, H.: Pointrcnn: 3d object proposal generation and
  detection from point cloud. In: CVPR (2019)

\bibitem{song2018pixeldefend}
Song, Y., Kim, T., Nowozin, S., Ermon, S., Kushman, N.: Pixeldefend: Leveraging
  generative models to understand and defend against adversarial examples. In:
  ICLR (2018)

\bibitem{song2018constructing}
Song, Y., Shu, R., Kushman, N., Ermon, S.: Constructing unrestricted
  adversarial examples with generative models. In: NeurIPS (2018)

\bibitem{stutz2019disentangling}
Stutz, D., Hein, M., Schiele, B.: Disentangling adversarial robustness and
  generalization. In: CVPR (2019)

\bibitem{su2018deeper}
Su, J.C., Gadelha, M., Wang, R., Maji, S.: A deeper look at 3d shape
  classifiers. In: ECCV Workshop (2018)

\bibitem{szegedy2013intriguing}
Szegedy, C., Zaremba, W., Sutskever, I., Bruna, J., Erhan, D., Goodfellow, I.,
  Fergus, R.: Intriguing properties of neural networks. ICLR  (2014)

\bibitem{tewari2017mofa}
Tewari, A., Zollhofer, M., Kim, H., Garrido, P., Bernard, F., Perez, P.,
  Theobalt, C.: Mofa: Model-based deep convolutional face autoencoder for
  unsupervised monocular reconstruction. In: ICCV (2017)

\bibitem{tung2017self}
Tung, H.Y., Tung, H.W., Yumer, E., Fragkiadaki, K.: Self-supervised learning of
  motion capture. In: NeurIPS (2017)

\bibitem{varol2018bodynet}
Varol, G., Ceylan, D., Russell, B., Yang, J., Yumer, E., Laptev, I., Schmid,
  C.: Bodynet: Volumetric inference of 3d human body shapes. In: ECCV (2018)

\bibitem{wang2018dynamic}
Wang, Y., Sun, Y., Liu, Z., Sarma, S.E., Bronstein, M.M., Solomon, J.M.:
  Dynamic graph cnn for learning on point clouds. ACM Transactions on Graphics
  (TOG)  (2019)

\bibitem{wu2017marrnet}
Wu, J., Wang, Y., Xue, T., Sun, X., Freeman, B., Tenenbaum, J.: Marrnet: 3d
  shape reconstruction via 2.5 d sketches. In: NeurIPS (2017)

\bibitem{wu2016learning}
Wu, J., Zhang, C., Xue, T., Freeman, B., Tenenbaum, J.: Learning a
  probabilistic latent space of object shapes via 3d generative-adversarial
  modeling. In: NeurIPS (2016)

\bibitem{wu20153d}
Wu, Z., Song, S., Khosla, A., Yu, F., Zhang, L., Tang, X., Xiao, J.: 3d
  shapenets: A deep representation for volumetric shapes. In: CVPR (2015)

\bibitem{xiang2019generating}
Xiang, C., Qi, C.R., Li, B.: Generating 3d adversarial point clouds. In: CVPR
  (2019)

\bibitem{xiao2018generating}
Xiao, C., Li, B., Zhu, J.Y., He, W., Liu, M., Song, D.: Generating adversarial
  examples with adversarial networks. In: IJCAI (2018)

\bibitem{xiao2019meshadv}
Xiao, C., Yang, D., Li, B., Deng, J., Liu, M.: Meshadv: Adversarial meshes for
  visual recognition. In: CVPR (2019)

\bibitem{xu2017feature}
Xu, W., Evans, D., Qi, Y.: Feature squeezing: Detecting adversarial examples in
  deep neural networks. arXiv preprint arXiv:1704.01155  (2017)

\bibitem{yan2016perspective}
Yan, X., Yang, J., Yumer, E., Guo, Y., Lee, H.: Perspective transformer nets:
  Learning single-view 3d object reconstruction without 3d supervision. In:
  NeurIPS (2016)

\bibitem{yan2018deep}
Yan, Z., Guo, Y., Zhang, C.: Deep defense: Training dnns with improved
  adversarial robustness. In: NeurIPS (2018)

\bibitem{yang2018pixor}
Yang, B., Luo, W., Urtasun, R.: Pixor: Real-time 3d object detection from point
  clouds. In: CVPR (2018)

\bibitem{yang2019pointflow}
Yang, G., Huang, X., Hao, Z., Liu, M.Y., Belongie, S., Hariharan, B.:
  Pointflow: 3d point cloud generation with continuous normalizing flows. In:
  ICCV (2019)

\bibitem{yang2019adversarial}
Yang, J., Zhang, Q., Fang, R., Ni, B., Liu, J., Tian, Q.: Adversarial attack
  and defense on point sets. arXiv preprint arXiv:1902.10899  (2019)

\bibitem{yang2018foldingnet}
Yang, Y., Feng, C., Shen, Y., Tian, D.: Foldingnet: Point cloud auto-encoder
  via deep grid deformation. In: CVPR (2018)

\bibitem{zhao2018natural}
Zhao, Z., Dua, D., Singh, S.: Generating natural adversarial examples. ICLR
  (2018)

\bibitem{zhou2019dup}
Zhou, H., Chen, K., Zhang, W., Fang, H., Zhou, W., Yu, N.: Dup-net: Denoiser
  and upsampler network for 3d adversarial point clouds defense. In: ICCV
  (2019)

\bibitem{zhou2018voxelnet}
Zhou, Y., Tuzel, O.: Voxelnet: End-to-end learning for point cloud based 3d
  object detection. In: CVPR (2018)

\bibitem{zuffi2019three}
Zuffi, S., Kanazawa, A., Berger-Wolf, T., Black, M.J.: Three-d safari: Learning
  to estimate zebra pose, shape, and texture from images" in the wild". In:
  ICCV (2019)

\end{thebibliography}
\end{document}